\documentclass{scrartcl}

\usepackage[english]{babel}

\usepackage[letterpaper,top=2cm,bottom=4cm,left=3cm,right=3cm,marginparwidth=1.75cm]{geometry}

\DeclareOldFontCommand{\sc}{\normalfont}{}

\setkomafont{captionlabel}{\bfseries}

\usepackage{amsmath}
\usepackage{amssymb}
\usepackage{graphicx}
\usepackage[colorlinks=true, allcolors=blue]{hyperref}
\usepackage[nameinlink,capitalise,noabbrev]{cleveref}
\usepackage[inline]{enumitem}

\usepackage{booktabs}
\usepackage{multirow, makecell}
\usepackage{array}
\usepackage[table]{xcolor}

\usepackage{siunitx}

\usepackage{todonotes}
\usepackage{subcaption}

\usepackage{microtype}

\usepackage{crossreftools}
\newcommand{\crefa}[2]{\hyperref[#1]{\cref*{#1}~(#2)}}
\newcommand{\nhparagraph}[0]{\vspace{0.5\baselineskip}\noindent}

\newcolumntype{C}[1]{>{\centering\arraybackslash} m{#1}}
\newcolumntype{R}[1]{>{\raggedleft\arraybackslash} m{#1}}
\newcolumntype{L}[1]{>{\raggedright\arraybackslash} m{#1}}

\crefname{RQ}{}{s}
\crefname{STITCHSEL}{Rule}{Rules}

\usepackage{csquotes}
\usepackage[sorting=none,style=numeric,maxbibnames=24,style=trad-abbrv,giveninits=true]{biblatex}
\title{Sharing Knowledge without Sharing Data: Stitches can improve ensembles of disjointly trained models}

\author{Arthur Guijt \and Dirk Thierens \and Ellen Kerkhof \and Jan Wiersma \and Tanja Alderliesten \and Peter A.N. Bosman}

\begin{document}
\maketitle

\begin{abstract}
Deep learning has been shown to be very capable at performing many real-world tasks. However, this performance is often dependent on the presence of large and varied datasets. In some settings, like in the medical domain, data is often fragmented across parties, and cannot be readily shared. While federated learning addresses this situation, it is a solution that requires synchronicity of parties training a single model together, exchanging information about model weights. We investigate how asynchronous collaboration, where only already trained models are shared (e.g. as part of a publication), affects performance, and propose to use stitching as a method for combining models.

Through taking a multi-objective perspective, where performance on each parties' data is viewed independently, we find that training solely on a single parties' data results in similar performance when merging with another parties' data, when considering performance on that single parties' data, while performance on other parties' data is notably worse. Moreover, while an ensemble of such individually trained networks generalizes better, performance on each parties' own dataset suffers. We find that combining intermediate representations in individually trained models with a well placed pair of stitching layers allows this performance to recover to a competitive degree while maintaining improved generalization, showing that asynchronous collaboration can yield competitive results. 
\end{abstract}


\section{Introduction}

In recent years, deep learning approaches have shown to be very capable of performing a wide variety of machine learning tasks. In fact, for many applications, like a wide variety of image analysis tasks, deep-learning approaches are often the state-of-the-art~\autocite{voulodimosDeepLearningComputer2018}. Many key advances in deep learning that led to the current state of affairs, have been made possible by large, public datasets. For example, ImageNet~\autocite{russakovskyImageNetLargeScale2015} (14M images, classification), MS COCO~\autocite{COCOCommonObjects,linMicrosoftCOCOCommon2014} (164K images, semantic segmentation), LAION~\autocite{schuhmannLAION400MOpenDataset2021} (400M images, captioning) are datasets of enormous size, each of which has been instrumental in achieving breakthroughs in performance.

Yet, large amounts of public data are not always available for every real-world task. For example, in the medical domain, this is often not the case because patient data cannot be published or shared - even with other institutes - without strict agreements and consent due to privacy concerns. Nevertheless, the ability to obtain the benefits associated with training on larger datasets, remains equally desirable.

\paragraph{Federated Learning}
One approach to avoid sharing raw data is federated learning.
At its core, federated learning is a process in which a model is collaboratively trained by multiple parties. Instead of sharing data, each party trains the model locally on their own data. During this process, trained models or model updates are occasionally exchanged, so that models can be combined. This combined model is then used by each party as the starting point of the next training iteration~\autocite{yangFederatedMachineLearning2019,shokriPrivacyPreservingDeepLearning2015}. This combination is often performed by averaging weights, which requires all  parties to use the same neural network architecture~\autocite{mcmahanCommunicationEfficientLearningDeep2017,zhangSecurePrivateHealthcare2023}. Typically, multiple rounds of local training, sharing and averaging are performed. While this approach avoids the need to share raw data, model updates need to be shared between parties. Furthermore, frequent sharing of model parameters is needed to achieve good performance~\autocite{zhangSecurePrivateHealthcare2023}. Hence, this approach requires compute clusters of different parties to be able to communicate automatically, which may not always be possible.

\paragraph{Asynchronous Learning}

In various settings, it could be advantageous if we could train neural networks independent of one another, then share or even publish them, and finally combine them. In many domains it is already common to publish code accompanied by trained network weights alongside papers, for example using repositories like \href{https://huggingface.co/models}{Huggingface's Model Hub}, in particular due to the increased need for openness in science. Examples also include popular large language models and generative models, for which only the models are shared, not the data (and oftentimes, the procedure) used to train them. Especially when these models become very large and are intended to be trained using very large datasets, training from scratch is especially costly and out of reach for most research institutes and most industries. Being able to reuse previous work is essential to enable improvements and research. When another party has additional data that could be useful, they could attempt to improve the network to suit their needs, reusing knowledge, without having to involve the original party to train on the original dataset again. This is also the underlying principle of modern foundation models~\autocite{Bommasani2021FoundationModels}.

While large models, and particularly foundation models, are an excellent resource, they do not always suffice and may not always be suitable for a specific task. Moreover, in the aforementioned scenario, where multiple parties wish to collaborate but are restricted from sharing their raw data and cannot establish an environment that supports synchronous federated learning, the setting is not well-suited for training a large foundation model. Instead, multiple models may have to be trained using smaller datasets at individual institutes. Note that this scenario also involves the situation in which multiple parties do not wish to -- or are not able to -- collaborate simultaneously. Instead, parties may join the collaboration at a later stage, either requesting access to existing models, or contributing models of their own. Although trained models may be exchanged more easily than raw data, a key question is, whether and how the knowledge embedded in such model can still be leveraged to improve performance for individual parties.


\nhparagraph{}
Within this work, we investigate the applicability and added value of approaches that utilize models trained on data from another party to improve the performance of a model trained on local data. We particularly study this topic in a real-world setting, using various real-world medical datasets for semantic segmentation. Key questions that we focus on are: (1) what performance improvement can be achieved by combining models compared to training on own data alone, and (2) how does the performance compare against approaches with more stringent requirements, like merging datasets - requiring training data to be shared - or federated learning - requiring active coordination between involved parties.
To utilize already trained models we consider various well-known approaches, such as ensembling, as well as a new method which adapts  stitching~\autocite{bansalRevisitingModelStitching2021,panStitchableNeuralNetworks2023,guijtExploringSearchSpace2024} for this task. The resulting technique not only combines the outputs of existing neural networks, but also integrates information encoded in hidden layers, even when the network architectures differ.

\section{Background}
Within this section we briefly introduce the application that we used as a case study (i.e., semantic segmentation), and discuss some approaches in literature that could be used to perform asynchronous learning.

\subsection{Semantic Segmentation}
The goal of semantic segmentation is to assign pixels (or voxels) in a given input image to one of the classes of interest, while pixels (or voxels) that do not belong to these classes are treated as background or ignored. In the medical domain, this process is referred to as delineation. It is commonly used to identify important anatomical structures relevant for diagnosis or treatment, such as organs or bony anatomy. 
For this task, neural networks have shown to be highly efficient and effective, both for 2D and 3D (medical) images, which we also consider in this work. Common architectures are the U-Net architecture~\autocite{ronnebergerUNetConvolutionalNetworks2015} and vision transformers like Swin UNETR~\autocite{hatamizadehSwinUNETRSwin2022}. Of particular interest is nnUNet~\autocite{isenseeNnUNetSelfconfiguringMethod2021}, which is a self-configuring pipeline based on the U-Net architecture that is easy to use and known to be capable of achieving state-of-the-art results for a wide range of (medical) image segmentation tasks.

\subsection{Approaches for Asynchronous Learning}
Various approaches exist that could be utilized for asynchronous learning, which we discuss below.
\paragraph{Fine-tuning} A neural network trained on one dataset can undergo further training on a different dataset, without re-initializing its parameters. Doing so naively, however, will likely lead to the destruction of the features already learned, a phenomenon known as \emph{catastrophic forgetting}~\autocite{frenchCatastrophicForgettingConnectionist1999,kirkpatrickOvercomingCatastrophicForgetting2017}.
In general, to preserve the features previously learned on another dataset, the network's final weights after further training need to remain close to the original network weights~\autocite{kirkpatrickOvercomingCatastrophicForgetting2017}. Correspondingly, using high learning rates or employing different procedures that induce large changes can hinder reuse of previously learned features. Therefore, in fine-tuning most network layers are often frozen, or low learning rates are used to attempt to preserve features and prevent large degradations in network performance. 
Even when configured successfully, fine-tuning can only leverage multiple datasets by training on these datasets sequentially. Hence, it cannot be used to combine independently trained networks from multiple parties.

\paragraph{Ensembles} A method to combine separate models, even if they have varying architectures, is to create an ensemble. An ensemble is a set of models of which the predictions are combined. Ensembles of deep neural networks, also called deep ensembles, are useful, due to their variance reducing and performance improving properties \autocite{wasayMotherNetsRapidDeep2020}. It is for this reason used by default in nnUNet~\autocite{isenseeNnUNetSelfconfiguringMethod2021} to combine model predictions, in particular those originating from different splits of the dataset, to obtain a single prediction corresponding to all folds. For a distributed approach, the ensemble offers key benefits: it is easy and inexpensive to construct, and requires nothing beyond sharing the final network with its corresponding weights. Key downsides to this approach are that the model size and computational costs are equal to the sum of the models being combined. Furthermore, each model is used separately as a black box. Models are unable to reuse knowledge from one another, only their final predictions are combined.


\paragraph{Weight-based Merging}
As discussed earlier, in the synchronous approach to federated learning it is common to average model weights. However, weight averaging only works if the networks share both identical architectures \emph{and} have \emph{weights} which are sufficiently similar. Even when the two networks have the same architecture, as to enable averaging, the weights can differ significantly, for example due to differing initializations. In such cases, averaging often yields a non-functional network. This limitation arises because of symmetries in weight space~\autocite{schafferCombinationsGeneticAlgorithms1992,thierensNonredundantGeneticCoding1996,stanleyEvolvingNeuralNetworks2002}. The same mathematical function can be represented in different ways within a neural network's parameters. Averaging weights without accounting for these symmetries can lead to the merging of unrelated features, thereby breaking the network's functionality. The most well-known example of such symmetry is the permutation symmetry in the hidden layers of a multilayer perceptron. This type of symmetry - or similar ones - can be accounted for and corrected by methods like `Git Re-Basin'~\autocite{ainsworthGitReBasinMerging2023}, Neuron Alignment~\autocite{uriotSafeCrossoverNeural2020}, and `ZipIt!'~\autocite{stoicaZipItMergingModels2023}. However, such methods can only be used to combine models if the networks are sufficiently similar in architecture, and if the symmetries of the layers used and their composition are known.

Recent work has shown that when models are derived from the same foundation model, for example through fine-tuning, they tend to remain sufficiently close in weight space. As a result, they can be combined back into such a large model using averaging, or other techniques~\autocite{goddardArceesMergeKitToolkit2024}. However, this still requires the involved parties to have used the same foundation model as a starting point, which is not the scenario we consider here where independently trained neural networks are shared.

\paragraph{Generative Models}\label{approach:generative-models}
An alternative to sharing models, is sharing \emph{synthetic} data generated using generative models trained on local data. If individual synthetic data records do not closely match any original raw data records, but the distributions of the synthetic and raw data are sufficiently similar, privacy concerns may be mitigated while key data characteristics are maintained. Neural networks may be fine-tuned using synthetic data from other parties, or a new network can be trained on all synthetic data, similar to a centralized data configuration~\autocite{chebykinHyperparameterFreeMedicalImage2024}. This approach has the advantage of being generically applicable, allowing any new neural network architecture to be trained. However, the synthetic data must still be generated and annotated by models. As such, it often deviates from the true data distribution, which can lead to similar issues to recursively training generative models on their own synthetic outputs, a process known to degrade performance notably \autocite{shumailovCurseRecursionTraining2024,shumailovAIModelsCollapse2024}.


\section{Stitch-ensembles}\label{sec:stitching}
The aforementioned permutation symmetry is not always present, and may not be the only type of symmetry. For instance, certain architectural choices, such as combinations of layers or specific activation functions, can introduce new symmetries. The \texttt{tanh} activation function, for example, is sign symmetric~\autocite{thierensNonredundantGeneticCoding1996}, whereas grouped convolutions, like those used in ResNeXt~\autocite{xieAggregatedResidualTransformations2017}, cannot be permuted freely. Instead of relying on predefined rules, corrections between representations can be learned by training an intermediate layer, as is done with stitching~\autocite{bansalRevisitingModelStitching2021}.

\paragraph{Stitching Neural Networks}
A general method for translating between two representations is to use a suitable layer or network, referred to as stitch, that is trained to perform this translation. This stitching approach can be used to assess representation similarities~\autocite{bansalRevisitingModelStitching2021,baloghHowNotStitch2024}, interpolate between various model scales~\autocite{panStitchableNeuralNetworks2023}, and define search spaces spanned by two networks for neural architecture search~\autocite{guijtExploringSearchSpace2024}. Because stitching transforms activations rather than weights, it can be used even when the underlying networks differ in both weights and architectural parameters, provided that a suitable translation mechanism (e.g., a layer or small network) is available~\autocite{panStitchableNeuralNetworks2023,guijtExploringSearchSpace2024}.

Stitching by itself does not specify how, or where the stitches should be introduced, as this is task- and network-dependent. In~\autocite{bansalRevisitingModelStitching2021}, the goal was to assess feature map similarity by stitching at wide variety of layer positions. This was expanded upon in~\autocite{baloghHowNotStitch2024}, by analyzing the impact of two training methodologies: task loss matching and direct matching. In~\autocite{panStitchableNeuralNetworks2023}, transformers of varying depths were stitched based on the location of the layer in the sequence, to interpolate between performance and computational cost trade-off of across different model scales. In~\autocite{guijtExploringSearchSpace2024} existing networks with differing architectures were converted into a directed acyclic graph (DAG), using a matching approach to select appropriate stitching positions. In all of these works, `simple' stitches consisting of a single layer were used.

In this work, we will perform stitching to neural networks generated by nnUNet~\autocite{isenseeNnUNetSelfconfiguringMethod2021}, which follow the UNet architecture~\autocite{ronnebergerUNetConvolutionalNetworks2015}. This architecture includes multiple parallel branches, and cannot be represented as a simple sequence of layers, applied one after the other, as it processes information at various spatial scales in parallel. Many architectures used for medical image segmentation share these characteristics. Therefore, stitching methods designed for strictly sequential architectures -- those architectures without parallel branches, as commonly assumed in existing work regarding stitching -- are insufficient for this setting.


\begin{figure}
    \centering%
    
    \subcaptionbox{%
    Given two networks: determine or select the pairs of layers where stitching will be applied. Here, we stitch between $v_A$ in network $A$ and $v_B$ in network $B$. Dashed lines indicate other networks; between which additional stitches may be considered.
        \label{diagram:stitching-1}%
    }{
        \includegraphics[width=\columnwidth]{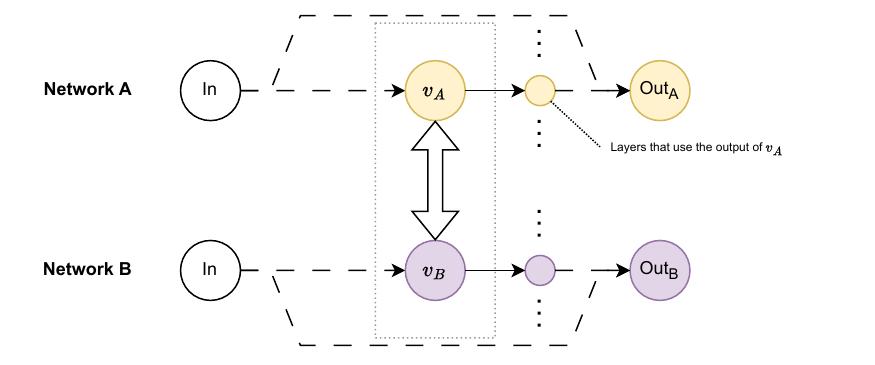}%
    }
    \subcaptionbox{%
    Combine the two networks by introducing stitching layers. The choice of whether to use the stitch is deferred by inserting a switch -- a multiplexer that selects between the original and the stitched connections.
        \label{diagram:stitching-2}%
    }{%
        \includegraphics[width=\columnwidth]{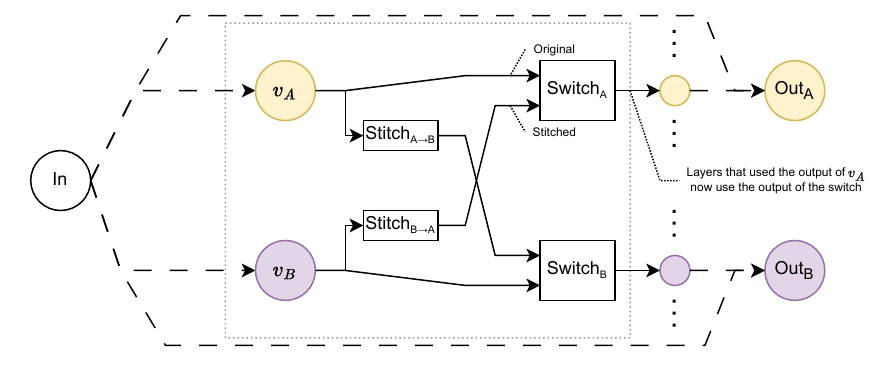}%
    }
    \caption{A schematic representation of stitching.}%
    \label{diagram:stitching}
\end{figure}
\begin{figure}
    \ContinuedFloat%
    \centering%
    
    \subcaptionbox{%
    Train the introduced stitching layers, for example using a direct matching loss, while keeping the original network layers frozen, such that they do not receive parameter updates. This includes setting layers to evaluation mode, to avoid batch statistics from being updated.
        \label{diagram:stitching-3}%
    }{%
        \includegraphics[width=\columnwidth]{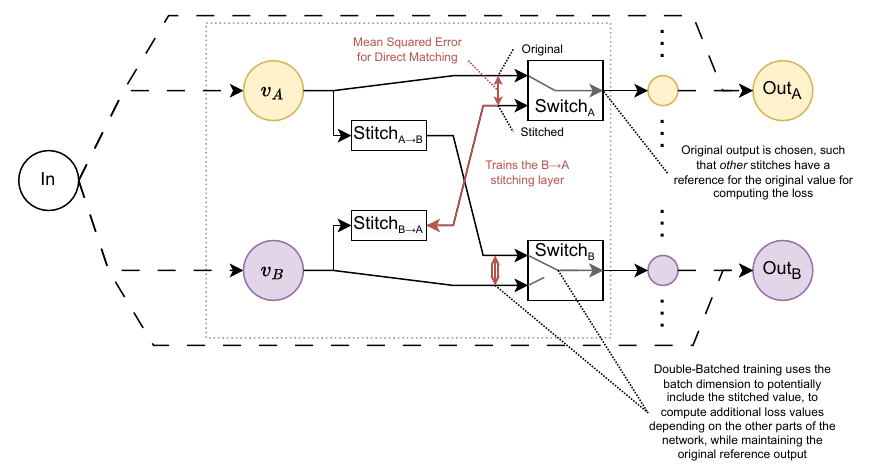}%
    }
    \subcaptionbox{%
    Select or combine representations at each switch to create a new, combined, network.
        \label{diagram:stitching-4}%
    }{%
        \includegraphics[width=\columnwidth]{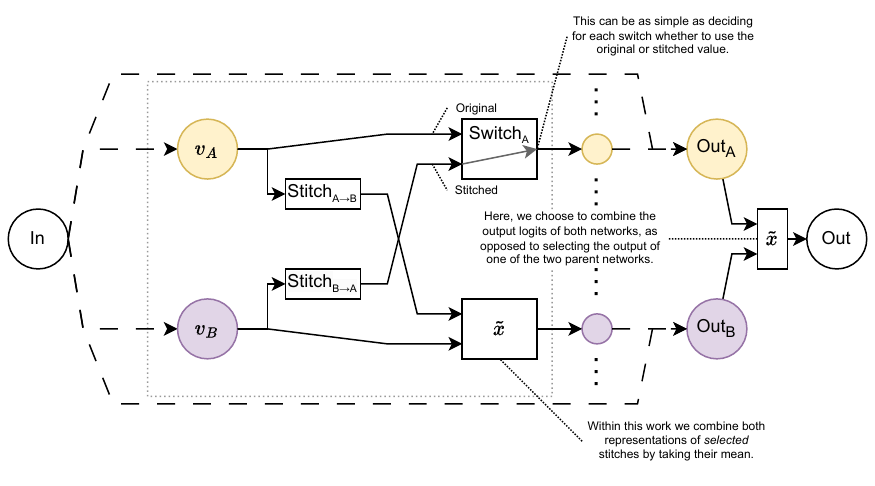}%
    }%
    \caption{A schematic representation of stitching.}%
\end{figure}

To address this, we build upon the approach proposed in~\autocite{guijtExploringSearchSpace2024}, in which neural networks are modeled as DAGs. In this formulation, each vertex represents an input, output or operator (layer) in the network, and each edge represents the flow of data from one layer to another. Each edge is annotated with the corresponding argument position or name, indicating how outputs are passed as inputs.

Stitching can be divided into several steps, which are shown in \cref{diagram:stitching}. First, as shown in \cref{diagram:stitching-1}, given two networks $A$ and $B$, represented as DAGs, a decision needs to be made about where to apply stitching. This involves selecting layer pairs that are considered potential stitching points. For example, one such pair of layers could be $v_A$ in network $A$ and $v_B$ in network $B$. Second, as shown in \cref{diagram:stitching-2}, the graph representations of networks $A$ and $B$ are combined, and selected layer pairs are modified. For each pair $v_A$ and $v_B$, stitching layers and switch layers are introduced, and all usages of the outputs of $v_A$ and $v_B$ are replaced with the output of the corresponding switch. Next, as illustrated in \cref{diagram:stitching-3}, the stitching layers are trained to transform the representation from the output of $v_A$ to resemble that of $v_B$. Finally, as shown in \cref{diagram:stitching-4}, the switches can be resolved by selecting either one of the inputs, or, as proposed in this work, by averaging them, to combine the feature maps. Each of the steps involves additional design choices and implementation considerations, which are discussed in detail below.


\paragraph{Where to stitch}
Before stitches can be introduced, as shown in \cref{diagram:stitching-1}, the locations between which stitches will be applied, must first be determined. Determining where to introduce stitches is key: layers whose features differ significantly cannot always be stitched without degrading performance~\autocite{bansalRevisitingModelStitching2021,baloghHowNotStitch2024}. Furthermore, if the goal is to reduce the computational cost of using a network, as in~\autocite{panStitchableNeuralNetworks2023,guijtExploringSearchSpace2024}, different stitching choices yield different trade-offs between this cost and performance.
Within~\autocite{panStitchableNeuralNetworks2023,guijtExploringSearchSpace2024} and in this work, multiple stitches are considered simultaneously to improve training efficiency. These multiple stitches are selected such that the resulting combined graph remains a DAG. Unlike~\autocite{guijtExploringSearchSpace2024}, the matching procedure here is modified to incorporate the approximate position of layers within the network, following \autocite{panStitchableNeuralNetworks2023}, and employs a modified version of Hirschberg's Algorithm~\autocite{hirschbergLinearSpaceAlgorithm1975} for matching. Unlike \autocite{panStitchableNeuralNetworks2023}, which introduces stitches unidirectionally (i.e., from network $A$ to $B$), our approach introduces stitches bidirectionally, allowing information to flow in both directions, i.e., from $A$ to $B$ \emph{and} from $B$ to $A$. Additionally, only layers operating at the same image scale are matched. Because all models are derived from highly similar architectures (specifically, nnUNet), the matching found in this work is mostly one-to-one, aligning layers at the same visual scale. Further details are provided in \cref{appendix:matching}. 

\paragraph{How to stitch}
In step 2, as shown in \cref{diagram:stitching-2}, the graph is transformed by introducing stitching layers. Technically, stitching layers can be any function that maps the input feature map to the output feature map. However, simpler layers, like a linear layer or a convolution with a small kernel, are preferred to minimize computational overhead \autocite{bansalRevisitingModelStitching2021}. More complex stitches may be necessary when feature representations differ, for example a resampling operation might be introduced if the visual scales between layers do not match.
In this work, all stitches use either a 1 $\times$ 1 convolutional layer or a linear layer, depending on the shape of the output tensor.

\paragraph{How to train a stitch}
As shown in \cref{diagram:stitching-3}, stitches must be trained to perform the desired transformation.
Various training methods have been proposed. In \autocite{bansalRevisitingModelStitching2021}, stitches are trained using a task specific loss and with the remainder of the network frozen. This approach is extended in \autocite{baloghHowNotStitch2024}, which evaluates the impact of the choice of training procedure when stitching is used to assess feature map similarity. Specifically, it compares direct matching -- training with mean-squared-error (MSE) between the stitched and original feature map -- and task specific losses.
In \autocite{panStitchableNeuralNetworks2023} a linear least squares fit between the original and target features is used to initialize a linear layer, effectively performing direct matching on a small subset of data points. Training then proceeds with a task-specific loss and the network unfrozen, with one random stitch trained per forward pass.
In \autocite{guijtExploringSearchSpace2024}, training is performed using direct matching, training multiple stitches in the same forward and backward pass, improving efficiency. 

\begin{figure}[tbhp]
    \centering%
    \includegraphics[width=0.38\columnwidth]{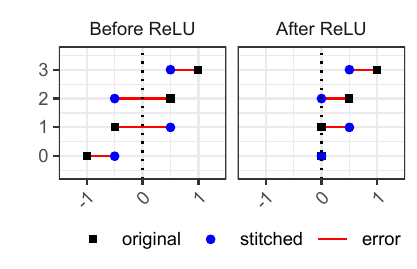}%
    \caption{
        The mean squared error between two values prior to and after a ReLU activation differs. Consequently, some differences are more important than others to the functioning of the neural network, while some can be ignored entirely.%
        \label{fig:relu-impact}%
    }
\end{figure}


The training procedure of stitches is key to their functioning. Poorly trained stitched, can easily disrupt any network that uses them. However, not all differences from the original input equally impact on downstream feature maps or, the final prediction. For example, differences among negative values are nullified when the ReLU activation function is applied, as illustrated in \cref{fig:relu-impact}. By computing the error at a later stage, more functionally important differences can be emphasized. While using a task-specific loss enables this, it also has notable drawbacks. The stitch must influence the network output to compute the loss, and if stitches are trained individually, only one stitch can be updated per forward pass, increasing computational cost. Additionally, in~\autocite{baloghHowNotStitch2024} demonstrates that stitches trained with task-specific losses are more prone to produce out-of-distribution activations.

We present a novel contribution -- double-batched training -- that enables stitches to be trained while accounting for later activations. This method also allows stitches to co-adapt to one another to reduce reconstruction errors, all within a single forward and backward pass. Further details are provided in \cref{appendix:double-batched-training}.

\paragraph{Ensembles using Stitches}
Finally, as shown in \cref{diagram:stitching-4}, we can extract new combined networks from the constructed stitched network. However, simply splicing networks by replacing connections -- as is commonly done in stitching -- does not necessarily improve performance. In this work, we propose combining both the stitched and original feature representation at one specific matching pair in the network by averaging, similar to an ensemble approach.

By combining outputs of multiple models, ensembles often provide better and more consistent predictions~\autocite{wasayMotherNetsRapidDeep2020}. Extending this idea, intermediate neural network representations -- which can encode high-level features like a car's wheel in a natural image~\autocite{olah2017feature} -- could also benefit from ensembling. A lack of robustness early in the network can propagate prediction errors at the output, but combining multiple networks at an intermediate feature level could help correct such mistakes in a semantically meaningful way. However,  while the output of a network has a predetermined meaning for a given task, and can be aggregated without much complexity, intermediate feature representations are subject to symmetries (similar to their weights), complicating direct combination. Stitching is used here to correct for these symmetries, enabling meaningful averaging of intermediate features.

\section{Experimental Setup}
First, to assess the new training method for stitches, this work investigates the following specific research questions:

\begin{enumerate}[label=\textbf{RQ \arabic*}, leftmargin=*]
    \item whether double-batched training improves the robustness and performance of the stitching approach, compared to conventional direct matching\label[RQ]{RQ:construct-better-network}.
\end{enumerate}

To address the two key questions introduced in the Introduction - (1) what is the performance improvement over training on own data alone, (2) how does the performance compare against approaches with more stringent requirements - this work investigates the following specific research questions:
\begin{enumerate}[label=\textbf{RQ \arabic*}, leftmargin=*, resume]
    \item how various forms of combining networks perform on a party's own data and compare to other collaborative approaches (federated learning, fine-tuning), considering practical factors such as hardware requirements and communication overhead\label[RQ]{RQ:comparison-combination}; and
    \item the impact of stitching on overall network performance\label[RQ]{RQ:impact-stitching}.
\end{enumerate}

Source code and result data is available at~\autocite{guijtSharingKnowledgeSharing2025}.

\subsection{Datasets}

In this section we describe the two datasets used within this work.  An overview of the individual datasets can be found in \cref{table:datasets-overview}.

\begin{table}
    \small
    \centering
    {\renewcommand{\arraystretch}{1.2}%
    \begin{tabular}{@{}R{0.21\linewidth}L{0.25\linewidth}L{0.25\linewidth}@{}}
        \toprule
         & \multicolumn{2}{c}{\textbf{Pair 1}}\\ \midrule
         & \multicolumn{1}{c}{\textbf{LUMC~\autocite{kostoulasConvolutionsTransformersTheir2023}}} & \multicolumn{1}{c}{\textbf{AUMC}} \\ \midrule
         & \multicolumn{2}{c}{\textit{Axial T2 Weighted MRI}} \\
        {\textit{\#samples}} & {185} & {68} \\
        {\textit{median resolution}} & {35 × 432 × 432} & {45 × 384 × 384} \\ 
        {\textit{median voxel spacing (mm)}} & {4.0 × 0.53 × 0.53 } & {3.3 × 0.52 × 0.52 } \\ 
        {\textit{median slice thickness (mm)}} & {4.0} & {3} \\ \bottomrule
        & & \\ \toprule
         & \multicolumn{2}{c}{\textbf{Pair 2}}\\ \midrule
         & \multicolumn{1}{c}{\textbf{HyperKvasir~\autocite{borgliHyperKvasirComprehensiveMulticlass2020}}} & \multicolumn{1}{c}{\textbf{CVC-ClinicDB~\autocite{bernalWMDOVAMapsAccurate2015}}}\\ \midrule
         & \multicolumn{2}{c}{\textit{Endoscope Video Frames (RGB)}} \\
         {\textit{\#samples}} & {1000} & {612} \\
         {\textit{median resolution}} & {530 × 621} & {288 × 384} \\
        \bottomrule
    \end{tabular}%
    }%
    \caption{Overview of datasets used within this work. Grouped by pair. While Pair 1 involves a 3-dimensional modality, 2D models are still used.}
    \label{table:datasets-overview}
\end{table}

\subsubsection{Organ-at-Risk Segmentation}\label{dataset:lumc-aumc}

We use a cervical cancer dataset from the radiation oncology department of the Leiden University Medical Center (LUMC)~\autocite{kostoulasConvolutionsTransformersTheir2023} consisting of 185 axial T2 weighted Magnetic Resonance Imaging (MRI) scans acquired for brachytherapy treatment planning between 2012 and 2020, of patients between the age of 23 and 90. If multiple scans from different scanners are available for a single patient, the scan acquired with the less common scanner is selected, as done in \autocite{chebykinHyperparameterFreeMedicalImage2024}. The following scanners were used: Philips Ingenia 1.5T (128 patients), Philips Intera 1.5T (36 patients), Philips Ingenia 3T (13 patients), Philips Achieva 3T (8 patients). Each scan corresponds to a unique cervical cancer patient within the LUMC dataset.
We use another cervical cancer dataset from the department of radiation oncology of the Amsterdam UMC (AUMC), comprising 68 axial T2 weighted MRI scans, acquired between 2015-2018, of patients between the age of 29 and 85, all acquired using the Philips Ingenia 3T. Each scan corresponds to a unique cervical cancer patient within the AUMC dataset.

Ethical approval was not required for this retrospective study, as determined by the medical research ethics committees of both participating institutions. Both committees confirmed that the Medical Research Involving Human Subjects Act does not apply, and therefore the study was exempt from formal ethical review. All data handling and processing were performed in accordance with local institutional guidelines and the General Data Protection Regulation (GDPR).

For our experiments these two datasets are primarily handled separately, emulating a distributed setting in which only trained models are shared. However, to establish a theoretical upper bound on performance if the data were pooled, as we are capable of doing so, we train a model on the combined dataset, as detailed in \cref{sec:tasks/alternativeapproaches}. Normalization and preprocessing are performed as automatically determined by nnUNet~\autocite{isenseeNnUNetSelfconfiguringMethod2021}.

The original delineations, made by clinicians in a clinical setting, namely for brachytherapy planning, include four regions of interest: bladder, bowel, rectum, and sigmoid. While an average of each metric over all regions of interest provides an indication of overall performance of the model, it can hide degradations to individual regions of interest when paired with an improvement to another region of interest. This is of particular concern as large regions of interest are generally less sensitive to errors with the metrics used within this article, making errors for smaller regions of interest have a larger impact on the overall score. As such, to keep evaluation and analysis simple, we focus on a single region of interest: the joint region of rectum and sigmoid. Due to large inter-observer variability in delineating the rectum and sigmoid individually -- largely because the boundary between these two regions is difficult to define in practice -- we have combined these two into a single joint region for delineation.


\subsubsection{Polyp}\label{dataset:polyp}
Similar to \autocite{chebykinHyperparameterFreeMedicalImage2024}, we use the CVC-ClinicDB~\autocite{bernalWMDOVAMapsAccurate2015} and the HyperKvasir~\autocite{borgliHyperKvasirComprehensiveMulticlass2020} datasets, each representing a different party. Both datasets consist of frames of gastrointestinal endoscopy videos with polyp annotations. The CVC-ClinicDB dataset consists of $612$ examples, whereas the HyperKvasir dataset consists of $1000$ examples. For CVC-ClinicDB, some annotated frames originate from the same source video. Using the provided video-to-frame mapping, we ensured that all frames from the same video were assigned to the same data split. This balancing of video sources prevents potential data leakage during cross-validation.

\subsubsection{Data Splitting}
A dataset of size $n$ is split into six parts of approximately equal size (i.e., $\left\lfloor \frac{n}{6}\right\rfloor$). Any remaining samples due to rounding are added to the last split, which is used as the test set. Note that $n$ differs between datasets. The remaining five parts are used to perform 5-fold cross-validation. Splitting is done randomly, except for the CVC-ClinicDB. As frames from the same video exhibit high similarity in CVC-ClinicDB, a bin covering algorithm, detailed below, was used to ensure frames from the same video remain within a single split, while still balancing the fold sizes.

The bin covering algorithm is a modified version of the bidirectional algorithm proposed in~\autocite{csirikTwoSimpleAlgorithms1999}. Similarly to this approach, it fills up the bins with the largest elements until the next video would cause the bin to contain more than a sixth of the frames. Unlike the original procedure, it defers the filling using the smaller elements until the largest elements have been distributed over the 6 bins. The smallest items (videos with the fewest frames) are then added to the bin with the fewest frames, until all videos have been distributed. This yielded a distribution of $[100, 100, 101, 101, 104]$ frames for each of the folds and $106$ frames for the test set dividing the $612$ frames total.

\subsection{Evaluation Metrics}\label{section:evaluation-metrics}

\paragraph{Dice}
The most commonly used metric for evaluating semantic segmentation models in medical imaging is the Dice Score ($\mathit{Dice}$), defined as:
\begin{equation}
    \mathit{Dice} = \frac{2 \mathit{TP}}{2 \mathit{TP} + \mathit{FP} + \mathit{FN}},
\end{equation}

\noindent where $\mathit{TP}$ denotes the true positives, $\mathit{FP}$ false positives, and $\mathit{FN}$ false negatives. Higher Dice scores indicate better performance. The true class refers to the region of interest, while the false class comprises the background class (and all other classes, if those were to be present). Each region of interest is generally only a small portion of an image, as such, the true class is usually (significantly) less common than the false class. The Dice Score is unaffected by solely classifying the background correctly: a trivial classifier predicting only the background yields a score of zero. This is unlike accuracy, which can be misleadingly high in the presence of class imbalance, especially for smaller structures.

The Dice Score's sensitivity to segmentation differences, such as over- and under-segmentation arising from inter-observer variation, depends on the size of the region-of-interest. The effect on the score depends on the size of the structure is, i.e., how large the $\mathit{TP}$ can maximally be, with larger regions being less sensitive to this effect. Given such variance, optimizing of the mean Dice score over all regions of interest tends to favor minimizing errors on smaller structures. As this complicates the analysis considerably, we have opted to focus on a binary segmentation task for both datasets above, simplifying the private datasets discussed in \cref{dataset:lumc-aumc}.

\paragraph{Hausdorff Distance}
Over-, or under-segmentation affect the Dice Score to a similar extent as delineating an unrelated region. Therefore, we additionally compute the 95th percentile Hausdorff Distance ($\mathit{HD95}$). Concretely, for every point on the predicted contour we compute the minimum Euclidean distance to the reference contour (and vice versa), then take the 95th percentile of those distances. HD95 thus measures the typical worst-case surface deviation while reducing sensitivity to extreme outliers; lower values indicate better geometric agreement.
For MRI scans in the LUMC and AUMC datasets the voxel spacing is accounted for to compute the deviation in \unit{\mm}. For the Endoscope Video Frames for HyperKvasir and CVC-ClinicDB unit spacing is assumed, as such the deviation is in pixels. Within each comparison the units are the same, and hence do not affect comparison between approaches.

\subsection{Network Training}
Networks for these datasets are created and trained using nnUNet~\autocite{isenseeNnUNetSelfconfiguringMethod2021}, with its standard configuration. We note that the network parameters selected by nnUNet differ across datasets for the same task, resulting in differing architectures for the Polyp task.

\subsection{Stitching}
While the networks in this work are trained using nnUNet, stitching requires a different, separate, training procedure. 

Stitches are trained using AdamW~\autocite{kingmaAdamMethodStochastic2017,loshchilovDecoupledWeightDecay2019} using a learning rate of $10^{-3}$, weight decay of $10^{-3}$ and a cosine annealing schedule without restarts~\autocite{loshchilovSGDRStochasticGradient2017}. The batch size requested from the dataloader is $5$ for the simple direct-matching approach, and $2$ for the double-batched approach. Note that the actual batch size doubles for the double-batched approach, as each sample is provided twice to the network, with the first used as a reference, and the second as the value to be optimized. The original approach trains for $30$ epochs, whereas the double-batched approach trains for $15$ epochs, to compensate for the cost induced by backpropagation through additional layers. Training all stitches took roughly $47$ minutes for the original approach and $54$ minutes for the double-batched approach for our initial pair of networks.

\subsection{Alternative Approaches}\label{sec:tasks/alternativeapproaches}
To evaluate the proposed approach, we compared to the following alternatives in addition to using stitching. We note that not all approaches assume the same operating conditions, for example, combining the datasets requires training data to be shared. A general overview of the approaches and their operating conditions, including stitching, can be found in \cref{table:approaches-overview}.

We exclude the use of synthetic datasets, for example those generated through generative models, as discussed in \autoref{approach:generative-models}. Creation of a synthetic dataset requires a party to prepare this data for the purpose of sharing. The availability of such a dataset could also influence other approaches, for example as addition to training data or to measure performance. We consider this to be outside of the scope of this work. 

\begin{table}
    \caption{Listing of approaches and their shorthands - as used in figures.}
    \label{table:approaches-overview}
    
    \centering
    
    {\renewcommand{\arraystretch}{1.2}%
    \begin{tabular}{@{}m{0.3\linewidth}C{0.2\linewidth}C{0.05\linewidth}C{0.05\linewidth}C{0.05\linewidth}C{0.05\linewidth}C{0.1\linewidth}@{}}
        \toprule
        \multirowcell{2}{\textbf{Approach description}} & \multirowcell{2}{\textbf{Approach}\\\textbf{(shorthand)}} & \multicolumn{2}{l}{\textbf{Dataset Use}} & \multicolumn{2}{l}{\textbf{Model Use}} & \multirow{2}{*}{\textbf{Online}} \\
         &  & \multicolumn{1}{c}{\textbf{a}} & \multicolumn{1}{c}{\textbf{b}} & \multicolumn{1}{c}{\textbf{a}} & \multicolumn{1}{c}{\textbf{b}} & \multicolumn{1}{c}{} \\\midrule
        Train on a & a & \checkmark &  &  &  &  \\
        Train on a, fine-tune on b & a → b &  & \checkmark & \checkmark &  &  \\
        Ensemble between models trained on a and b individually & a \& b &  &  & \checkmark & \checkmark &  \\
        Ensemble (with fine-tuning) & a \& b → a & \checkmark &  & \checkmark & \checkmark &  \\
        Federated learning & a $\upuparrows$ b & \checkmark $^{1}$ & \checkmark $^{1}$ & \checkmark & \checkmark & \checkmark \\
        Merge datasets & a + b & \checkmark & \checkmark &  &  &  \\
        Stitching models trained on a and b individually using dataset c & a ←[c]→ b & \multicolumn{2}{c}{\checkmark $^2$} & \checkmark & \checkmark & \multicolumn{1}{l}{} \\
        Stitching models trained on a individually and b finetuned on a using dataset a & a ←[a]→ b → a & \checkmark & & \checkmark & \checkmark & \multicolumn{1}{l}{} \\ \bottomrule
    \end{tabular}
    }%
    \vspace*{0.2em}
    \setlist{nolistsep}
    \begin{enumerate}[noitemsep]
    \item[$^{1}$] Datasets need to be present on different machines, but do not need to be shared.
    \item[$^{2}$] Either dataset a or b may be used for dataset c.
    \end{enumerate}
\end{table}

\paragraph{Basis Networks and their Ensembles}
The first alternatives are the basis networks, i.e., the networks trained on individual datasets and their ensemble.
The basis networks are created using nnUNet~\autocite{isenseeNnUNetSelfconfiguringMethod2021}, which determines both the architecture and the training procedure of these networks automatically.
The ensemble aggregates the predictions of each model matching the same fold index, e.g., model A-fold-0 and model B-fold-0, trained on datasets A and B respectively.

\paragraph{Fine-tuning}
Beyond creating an ensemble, a neural network trained on another party's local data can be further trained using local data.

We evaluate a simple configuration that minimizes the risk of catastrophic forgetting by using the smallest learning rate that is a power of $10$ that made progress during training, to avoid large weight perturbations during training.
We continue training the network with a smaller base learning rate of $10^{-5}$ for $1000$ epochs. In addition, the learning rate has a warm-up phase, increasing the learning rate by linearly interpolating between the final non-zero learning rate and the current learning rate of the original (polynomial) schedule for the first $333$ epochs. This alternate schedule is introduced to alleviate issues caused by a dataset shift.

As fine-tuning does not preclude the creation of an ensemble, we additionally evaluate an ensemble between the network trained on local data and the fine-tuned network.

\paragraph{Federated Learning}
This configuration trains a network jointly with other parties in lockstep. After a certain number of training batches, at the end of every epoch, each party shares their weights with all other parties (or a central node) and waits for the others to share their network weights, so that the weights can be averaged. Training then resumes starting with the new (averaged) weights.

Within this work, we implemented federated learning by modifying nnUNet. Our implementation uses the combined dataset's architectural parameters, as determined by nnUNet, assuming a real distributed implementation could share or combine dataset fingerprints. Training is performed by using Ray~\autocite{rayParallelLibrary} to coordinate multiple individual nnUNet training procedures -- one per dataset representing a party. Once an epoch has completed, weights are sent to a central actor for averaging, which pauses training on a specific dataset until all weights have been received, averaged, and replaced.

\paragraph{Merge Datasets}
This is the network that is obtained by training on all data pooled together. This configuration provides us with knowledge on what further performance could be obtained through sharing data, or alternatively, how much performance is lost by circumventing data sharing. While difficult, it is not impossible to create public datasets. For example, the Medical Segmentation Decathlon~\autocite{simpsonLargeAnnotatedMedical2019,antonelliMedicalSegmentationDecathlon2022} published 2633 scans across a variety of tasks. In \autocite{simpsonLargeAnnotatedMedical2019} they note that privacy considerations, especially with regard to protected health information (PHI), are a notable challenge for letting institutes share data, given the time and monetary cost associated with publishing such data.
Like the basis networks, this network is trained using nnUNet~\autocite{isenseeNnUNetSelfconfiguringMethod2021} with no further modifications.

\subsection{Hardware}
All networks are trained on a server with {2x Intel(R) Xeon(R) Bronze 3206R CPU @ 1.90GHz} and 3x {NVIDIA RTX A5000} GPUs. Each training run utilizes a single GPU, except for federated learning, which uses one GPU per party involved. Each machine has 93GB of RAM available, sufficient to load all training data into memory.

\subsection{Experiments}
Below, we discuss the two sets of experiments performed. The first set compares the impact of the new training methodology, while the second set analyzes the usefulness of stitching and averaging the resulting representations.

\subsubsection{Stitch Training Comparison}
The result of training and stitching two networks is a large neural network augmented with additional stitches and corresponding switches. Each switch can be set to use either the stitch or the original input.

We first verify that double-batched training improves the robustness of the stitches compared to training using only direct matching. To do this, we sample a variety of networks from the larger stitched network. For each count $c$ between 0 and the total number of stitches $\ell$ in total, we generate three networks for which $c$ switch layers (excluding the output switch) are selected to use the stitched input as their output, uniformly at random without replacement. Note that not all selected stitches are necessarily active, as the output of a switch may be unused itself due to another stitch or switch. The output is selected at random between the two parent networks. Duplicate network configurations are removed prior to evaluation. We perform this analysis using only the first fold of stitching the AUMC and LUMC networks, stitched using LUMC data, and evaluate the resulting networks on the corresponding LUMC validation set.

\subsubsection{Network Performance}

\begin{figure}
    \centering%
    \includegraphics[width=\columnwidth]{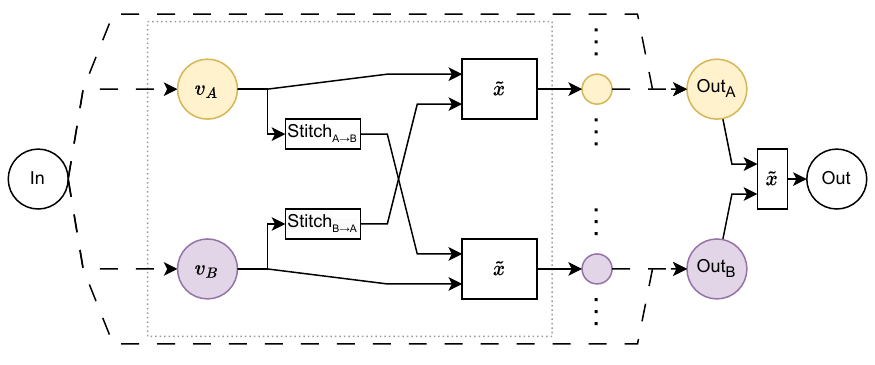}%
    \caption{When using stitching for combining networks, rather than selecting an intermediate representation, we use the average of the original and stitched representation instead. This was done by modifying each switch, including a third, averaged option between the two input feature maps.}
    \label{diagram:stitch-switch-with-average}
\end{figure}

\paragraph{Performance for Alternative Approaches}
Before we continue evaluating stitching, specifically, we first run and evaluate the baselines, i.e., the approaches in \autoref{table:approaches-overview} that do not involve stitching. In order to give a general overview of the baselines, and how they perform, we will utilize a multi-objective plot. Where  each axis represents the performance on one of the two datasets (objectives) individually, providing insight between the trade-off between the two objectives of various approaches. Here, we choose to showcase each of the folds individually, as to showcase the variance between folds. For brevity, metrics will be averaged over folds from this point onwards.

\paragraph{Stitch Selection} Stitching produces a network with different performance for each combination of stitches. Unlike other approaches, which generally produce a single network as output, stitching can produce many different networks with relatively little additional effort after the initial training procedure. For practical reasons, we wish to select a single network here.

Furthermore, as~\autocite{guijtExploringSearchSpace2024} indicated that searching through the entire space is difficult, we instead enumerate a subset of stitched networks. These networks form an ensemble of both input networks, enhanced by using the two stitches belonging to the same matched pair, and averaging their outputs. Specifically, we stitch from $v_A$ to the output of $v_B$, which is then averaged with the output of $v_B$, as shown schematically in \cref{diagram:stitch-switch-with-average}, and similarly from $v_B$ to $v_A$. We note that these two stitches combine well, as the computations they depend on and those they affect are independent.

While we wish to know how well these networks perform, here we deliberately avoid evaluating all networks on the test set and use validation data instead. A network should not be selected based on test-set performance, as this would unduly bias stitching as an approach on the test set.

Furthermore, when evaluating based on validation performance, perspective matters. Since stitching the AUMC and LUMC networks using LUMC data requires access to LUMC data, we are operating from LUMC's perspective. Accordingly, the validation score used corresponds to the LUMC validation set (unless defined otherwise). Performance on the AUMC dataset would remain unknown unless the model was shared with AUMC for evaluation. 

As an alternative to selecting stitches based solely on performance, we also select based on the position of the stitch itself. If a trend emerges indicating where performance tends to be the highest, such a pattern could serve as a general heuristic.

We consider selecting stitches according to one of the following strategies:

\begin{enumerate}
    \item Lowest mean rank (given current fold) \label[STITCHSEL]{stitch-sel:simple}
    \item Lowest mean rank over all folds \label[STITCHSEL]{stitch-sel:accumulate}
    \item Overall best position: Stitch index with the best performance across all networks and datasets. \label[STITCHSEL]{stitch-sel:a-priori-rule}
\end{enumerate}

The summary metric used to rank a stitch is the mean of its ranking based on the Dice Score, and its ranking based on HD95, relative to other stitches from the same network and evaluated on the same dataset. Both rankings are oriented such that lower is better. No aggregation of Dice Score or HD95 needs to take place, as we only segment a single region of interest for both datasets.

Strategy~\ref{stitch-sel:simple} greedily chooses the best stitch independently for each fold. Given the number of stitches evaluated, the selected network may achieve high performance in part due to random chance, potentially overfitting to its corresponding validation set. As architectures have been kept fixed across folds, we can consider the performance of the same stitch across different folds. Selecting a stitch based on mean performance over all folds could help mitigate this issue. This is the goal of strategy~\ref{stitch-sel:accumulate}. Strategies \ref{stitch-sel:simple} and \ref{stitch-sel:accumulate} use only the validation score from their respective (local) datasets.

If there is a trend based on the location of the stitches, the position itself serves as a useful criterion. Strategy~\ref{stitch-sel:a-priori-rule} analyzes the general performance across all networks, and provides a rule of thumb for which stitch to evaluate based on the overall performance obtained when stitching at that position. Although, strategy~\ref{stitch-sel:a-priori-rule} uses information that is typically unavailable in practice, it offers a useful guideline that could reduce the need to train and evaluate all stitches, provided such a positional trend exists.

To visualise a potential positional trend, we plot the stitch index, which increases along the depth of the network, against the performance metrics calculated on the validation set. For brevity and clarity, plotted values represent averages over the folds.

\paragraph{Multi-Dataset Overall Comparison}
First, we will perform an overall comparison across datasets by taking a multi-objective perspective. This perspective provides insight whether stitching has provided a generalist network that works well on both datasets, or whether the result is specialized on a single party's dataset. For brevity, plotted values represent averages over the folds. This view provides a good overview, but in practice we cannot select based on this plot, as we do not have access to the dataset of another party.

\paragraph{Own-Data Performance Comparison}
While we would like to obtain a general model, in practice performance on own data is the only metric we readily have access to.

We perform each experiment on both pairs of datasets, including experiments from the perspective of each party in the dataset. First, all baselines and corresponding networks are prepared and trained on both dataset pairs, then evaluated on the validation sets of each fold, as well as on the test sets.

We perform two-sided Wilcoxon signed rank tests, paired by the corresponding sample in the test set and the fold the model was trained on, with Holm-Bonferroni correction applied for multiple comparisons~\autocite{holmSimpleSequentiallyRejective1979}. Tests are performed separately for each perspective and restricted to configurations feasible from that perspective. For example, we only consider fine-tuning one's own data, or stitching using one's own data. The core question addressed here relates to \cref{RQ:comparison-combination}: whether some form of combination is beneficial for a single party. Accordingly, the test compares the configuration trained solely on own data, denoted as either a or b, with other approaches that involve some form of combination, all evaluated using only data of the party's own dataset.

We showcase the performance distribution on the given dataset over all patients and folds, grouped by the approach that produced the plots, using a combined violin boxplot.

\section{Results}
In this section, we discuss the results of the experiments described previously. First, we validate whether the improvements to the stitching method have enhanced stitch performance. Second, we present results obtained using the alternative approaches. Finally, we discuss the results from stitching, and their implications.

\subsection{Stitch Training Comparison}
\begin{figure}
    \centering
    \includegraphics[scale=0.67]{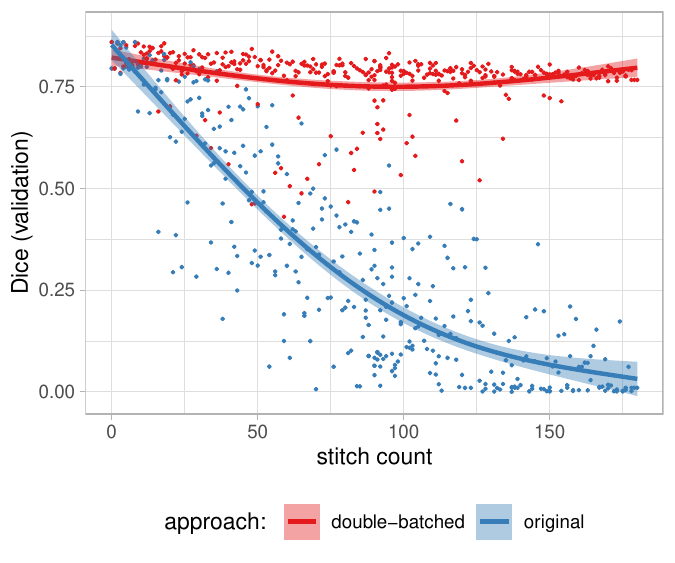}%
    \vspace{-0.7em}%
    \caption{Performance of a random sample of networks with varying numbers of stitches selected is shown. As the number of stitches increases, the performance of sampled networks degrades when stitches are trained with the original direct-matching approach (blue). The new double-batched approach (red) maintains higher performance, with most networks performing close to that of the basis networks used.}
    \label{fig:stitch-count-performance}
\end{figure}

In \cref{fig:stitch-count-performance}, we compare the performance of networks trained using direct matching versus double-batched training, evaluated over an increasing number of stitches. The actual number of \emph{active} stitches is lower, as some stitch outputs may be unused -- for example, when another stitch supersedes it. When trained using only direct matching, performance degrades significantly as more stitches are added. Double-batched training substantially mitigates this degradation without the need to train each stitch individually using a task-based loss. The resulting stitches are more robust, with most sampled networks maintaining performance levels close to their parent networks. However, performance at low stitch counts is very similar between methods, suggesting that the current configuration primarily improves co-adaptation of multiple stitches, rather than the performance of individual stitches alone. For the remainder of this work we employ the double-batched training procedure for stitches due to its superior performance observed here, even though no clear advantage is evident when only using a small number of stitches.


\subsection{Network Performance}
In this section we discuss the effect of stitching and alternative techniques for combining models.

\begin{figure}
    \centering
    \subcaptionbox{%
        Performance for dataset pair (a) LUMC \& (b) AUMC
    }{%
    \includegraphics[scale=0.67]{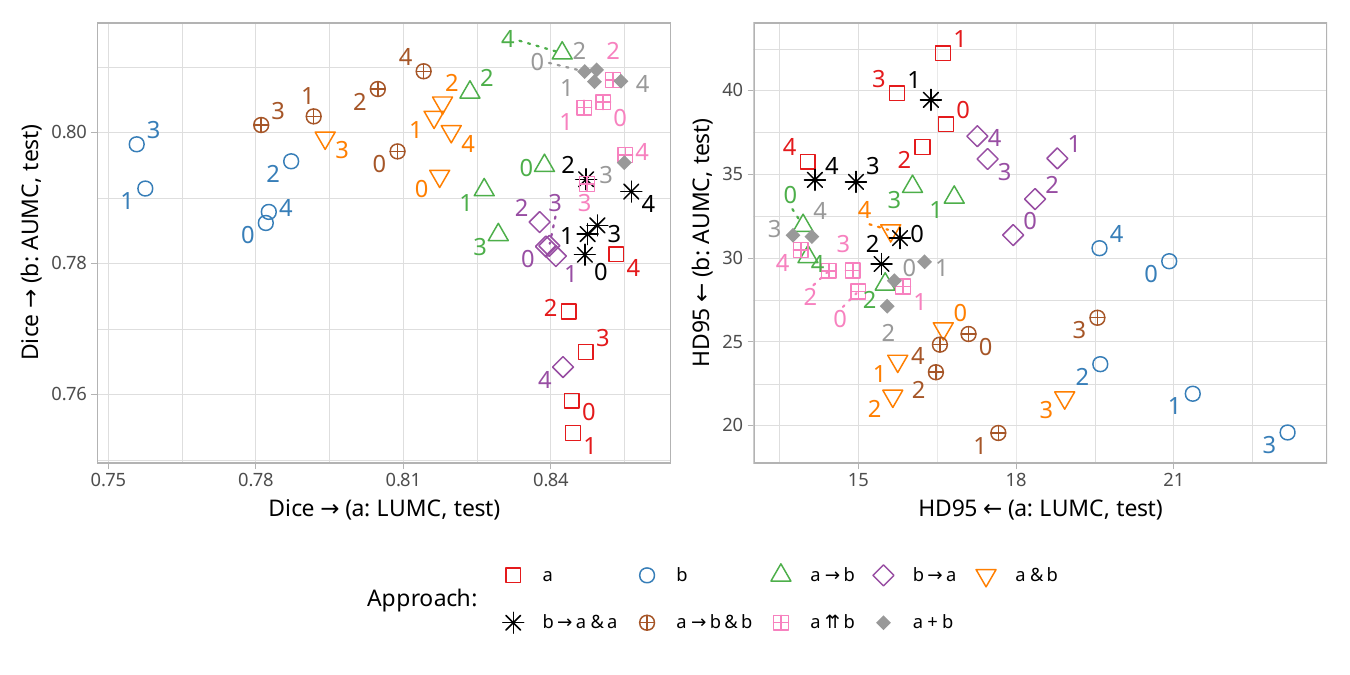}%
    \vspace{-0.7em}%
    }\\%
    \subcaptionbox{%
        Performance for dataset pair (a) HyperKvasir \& (b) CVC-ClinicDB
    }{%
    \includegraphics[scale=0.67]{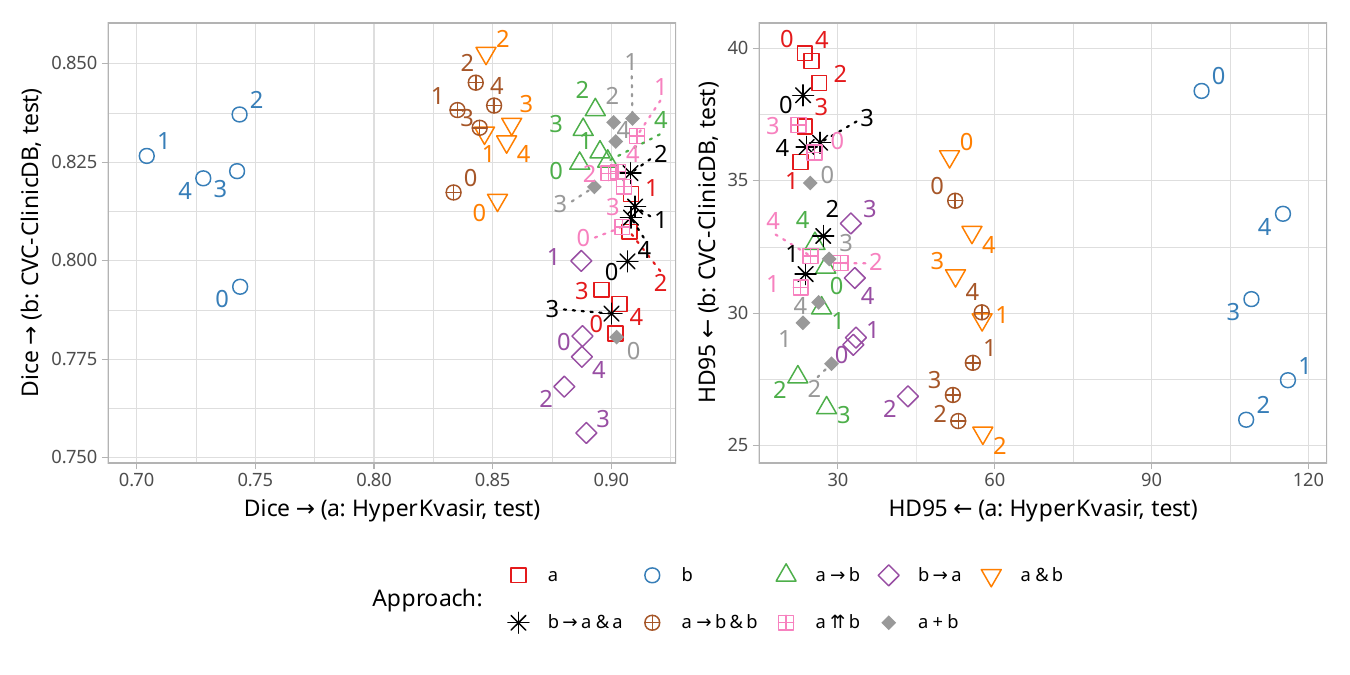}%
    \vspace{-0.7em}%
    }%
    \caption{Performance of baselines, all folds are shown and annotated by their fold index.}
    \label{fig:alternative-approach-performance}
\end{figure}

\paragraph{Performance for Alternative Approaches}
In \cref{fig:alternative-approach-performance} a bi-objective plot for both pairs of datasets is shown. Each point represents a model annotating a region of interest, the joint region of rectum and sigmoid for the LUMC and AUMC datasets, and polyp for CVC-ClinicDB and HyperKvasir. Networks trained on a single dataset, like `a' (red square) and `b' (blue circle) perform well on their own data, but do not generalize as well to the other dataset. The improvement obtained by combining datasets depends on the quantity and quality of the data originally available. Parties with smaller datasets and less varied data experience larger increases on the other party's dataset, and stand to benefit the most from using external data. In particular, the gap in performance for models trained solely using the CVC-ClinicDB dataset is large: on the HyperKvasir test set training on HyperKvasir alone is roughly $0.15$ DC points and $80$ HD95 points, better; while the converse, between HyperKvasir and CVC-ClinicDB on CVC-ClinicDB's test set, is $0.03$ DC points and $5$ HD95 points. 

Consistency is key, and any form of combining networks results in a considerable improvement in robustness and overall performance on both datasets compared to the networks derived from a single network. Furthermore, the variance between folds is generally smaller. Both combining datasets (gray rombus) and federated learning (pink crossed rectangle) generally result in highly desirable trade-offs, with their performance metrics ending up near the top right corner for Dice Score, and near the bottom left corner for HD95.

The impact of fine-tuning is much more variable, while it can yield performance comparable or exceeding to these approaches, performance cannot be guaranteed to be better. For example, while fine-tuning the LUMC network with AUMC data (green triangle) yields better or equivalent performance on both test sets when compared to training only on AUMC data (blue circle), this is not the case when fine-tuning the network trained on AUMC network with LUMC data (purple rombus), which outperformed on the LUMC test set by the network which has only been trained on LUMC data.

Improving these results, for example by further tuning the parameters, is complex, as each party only has access to their own local data. For example, if nnUNet's default learning rate of $10^{-2}$ is used, significant perturbations to the weights occur, causing catastrophic forgetting: the resulting model performs similarly to training from scratch on local data. While this may result in favorable performance on local data, and hence be a reasonable learning rate when tuning based on own local data, this is generally detrimental for performance on the other party's dataset. Blindly maximizing performance on local data does therefore not necessarily result in a good all-round model.

For the ensemble (orange downward triangle) the effect is not uniform across datasets. As noted previously, the constituent models have notable differences in performance. Let \emph{a} represent the party with the larger dataset of the pair and \emph{b} the other party, then a model trained on \emph{b}'s data performs much worse than a model trained with \emph{a}'s data on the \emph{a}'s test set. This difference is much larger than the difference between the model trained on \emph{a}'s and \emph{b}'s data, evaluated on \emph{b}'s test set. When aggregating predictions, this leads to a skew in performance of the resulting model, with performance on \emph{b}'s data improving, while the performance on \emph{a}'s is degraded due to the bad quality predictions of the model trained solely with \emph{b}'s data.


\begin{figure}
    \centering
    \subcaptionbox{%
        Performance for dataset pair (a) LUMC \& (b) AUMC
    }{%
    \includegraphics[scale=0.67]{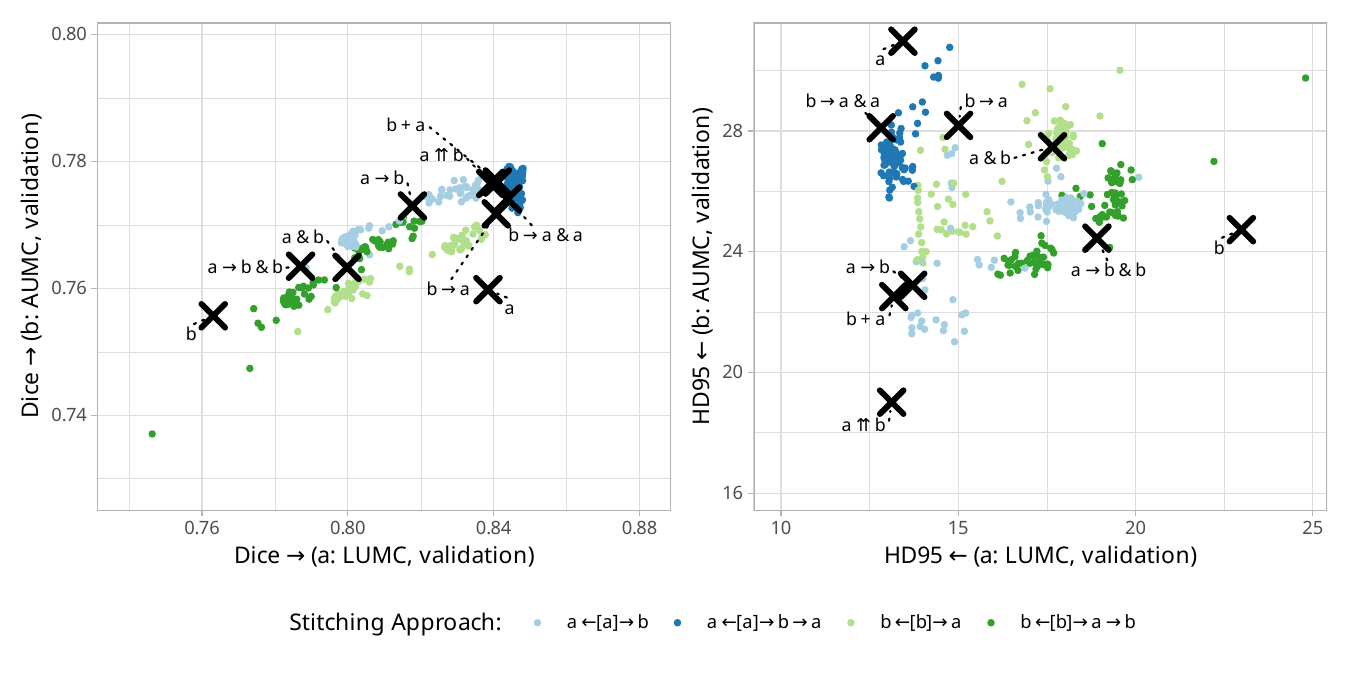}%
    \vspace{-0.7em}%
    }\\%
    \subcaptionbox{%
        Performance for dataset pair (a) HyperKvasir \& (b) CVC-ClinicDB
    }{%
    \includegraphics[scale=0.67]{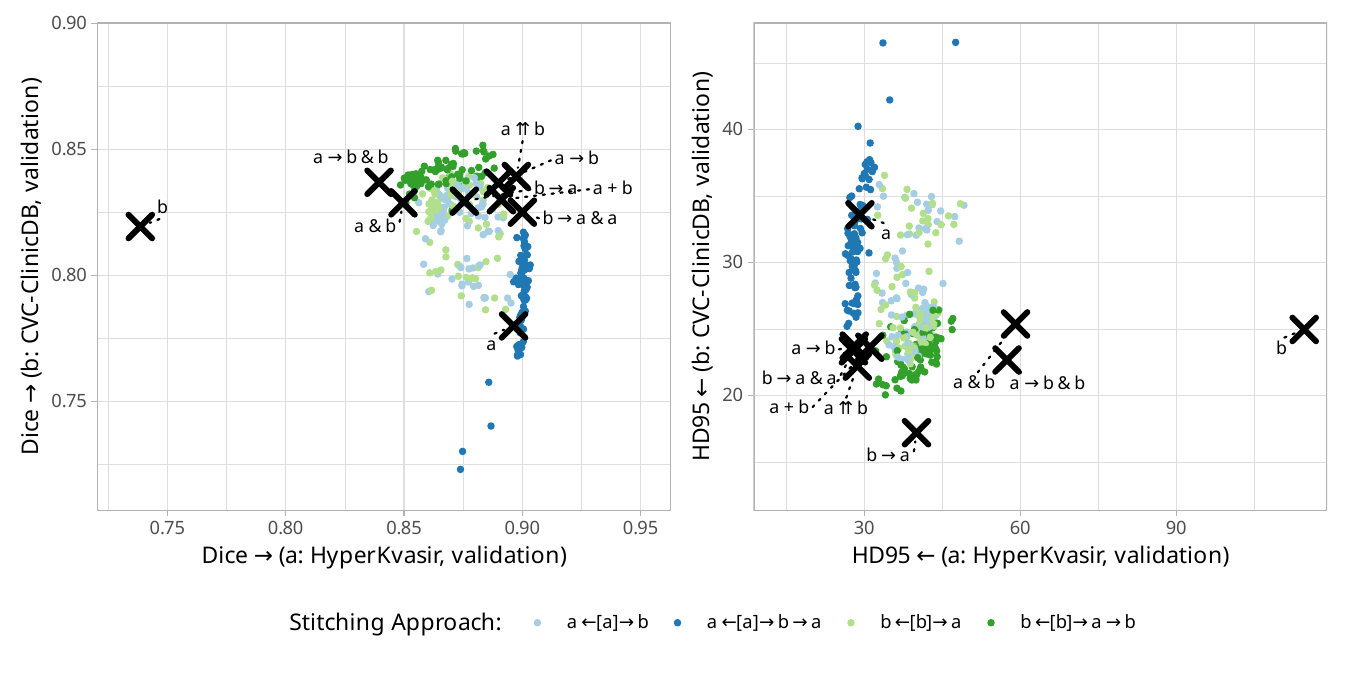}%
    \vspace{-0.7em}%
    }%
    \caption{Performance of stitches and baselines on validation, mean was taken over folds, stitches are aggregated by stitch index.}
    \label{fig:stitch-validation-performance}
\end{figure}

\paragraph{Stitch Selection - Is there a trend between position and stitch performance?}
In \cref{fig:stitch-validation-performance}, the mean performance of all folds on the \emph{validation} set for each dataset is shown. Visually, many of the stitched networks fill the gap between the reference approaches. The key difference between the ensemble 'a \& b' and a stitched network following 'a ←[a]→ b' is the inclusion of a stitch and averaging operation. The inclusion of the right stitch can improve the corresponding ensemble notably, potentially to the degree of performing similarly to sharing and merging datasets or federated learning, while the worst choice can result in degraded performance.

\begin{figure}
    \centering
    \subcaptionbox{%
        Performance for dataset pair (a) LUMC \& (b) AUMC
    }{%
    \includegraphics[scale=0.67]{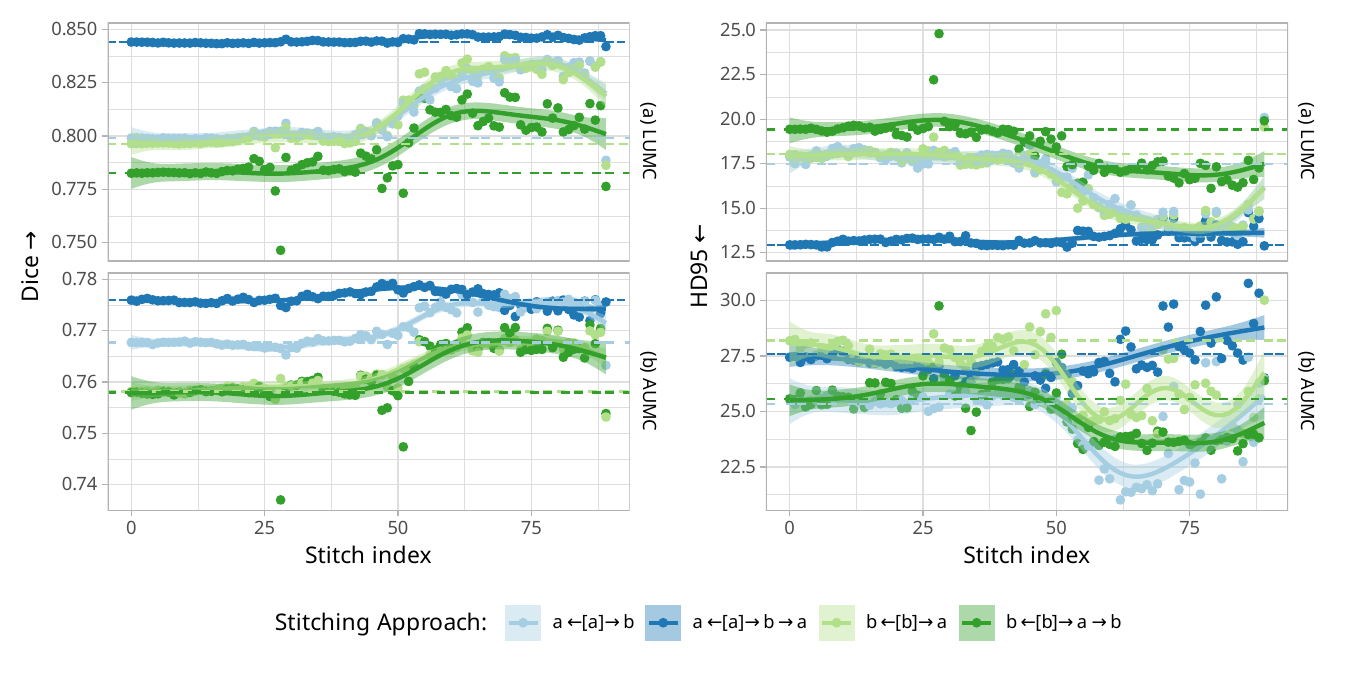}%
    \vspace{-0.7em}%
    }\\%
    \subcaptionbox{%
        Performance for dataset pair (a) HyperKvasir \& (b) CVC-ClinicDB
    }{%
    \includegraphics[scale=0.67]{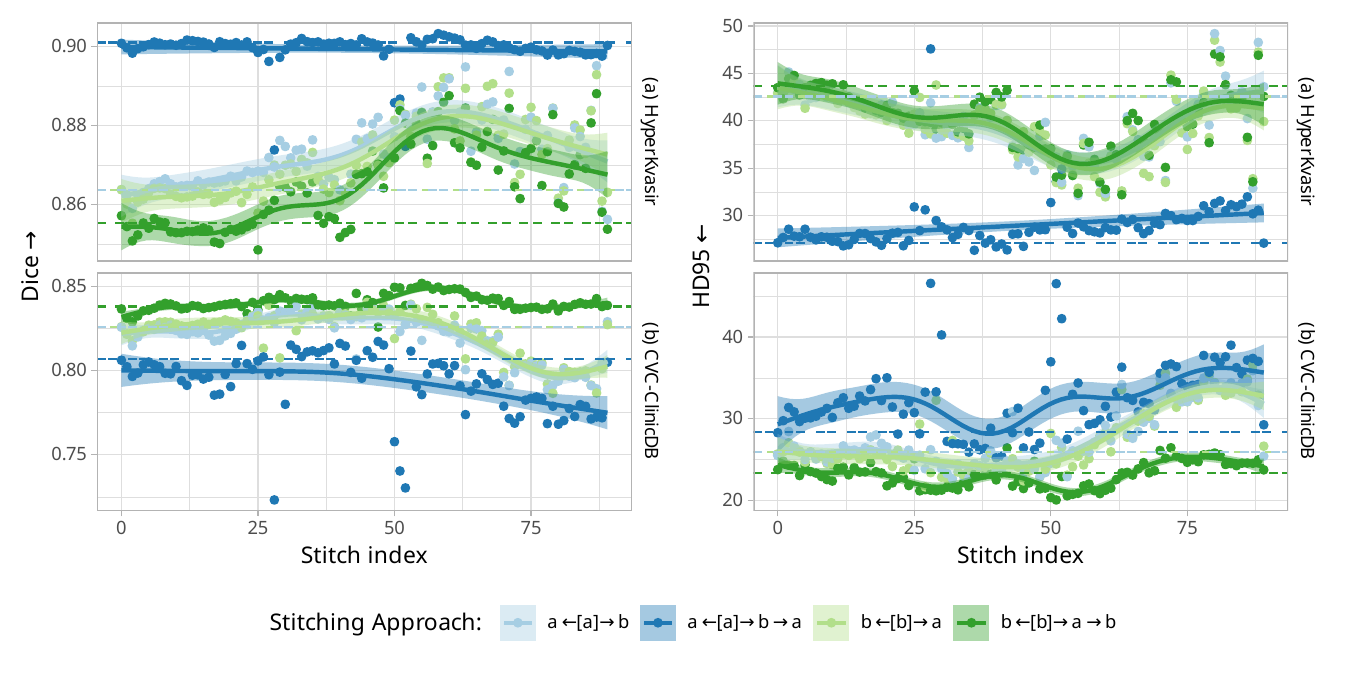}%
    \vspace{-0.7em}%
    }
    \caption{Performance of networks on validation data, averaged over folds by stitch index. Stitch indices are in order of the topological sorting of the network, and correspond to the `progress' along the network. Dashed line is the performance of the network without any added stitches, i.e., an ensemble with shared preprocessing.}
    \label{fig:stitch-validation-performance-positional}
\end{figure}

In \cref{fig:stitch-validation-performance-positional}, we show the performance based on where stitches are inserted, where higher indices correspond to stitching later in the network. In this figure the dashed line represents the stitched network with no added stitches. The difference between this line, and the networks including stitches, shows that the chosen stitch is a key driver for differences in performance. Furthermore, positional trends appear within the data. For example, it can be seen that early stitches tend to have little effect, and are generally quite similar to not introducing a stitch.

For most configurations and datasets we observe a performance improvement when stitching the latter part of the network, typically between indices 50 and 65. In the bi-objective plots, performance on the AUMC and LUMC datasets appears correlated, whereas this correlation is absent for HyperKvasir and CVC-ClinicDB datasets. Notably, the results on the CVC-ClinicDB dataset do not improve when stitching later in the network, unlike other datasets.

Overall, stitches 54, 58 and 59 perform the best with most nearby stitches not far behind. As such, for these networks \cref{stitch-sel:a-priori-rule} would indicate selecting stitch 54, approximately $60.7\%$ along the network.  \cref{stitch-sel:simple} and \cref{stitch-sel:accumulate}, both -- except for one fold -- prefers later stitches on the LUMC and AUMC datasets. For HyperKvasir and CVC-ClinicDB, the preferred stitch position varies: \cref{stitch-sel:simple} sometimes selects a stitch early in the network, whereas \cref{stitch-sel:accumulate} generally favors later stitches. Stitching is generally not beneficial for the first 20 positions, and the last stitch consistently performs the worst.

\begin{figure}
    \centering
    \subcaptionbox{%
        Performance for dataset pair (a) LUMC \& (b) AUMC
    }{%
    \includegraphics[scale=0.67]{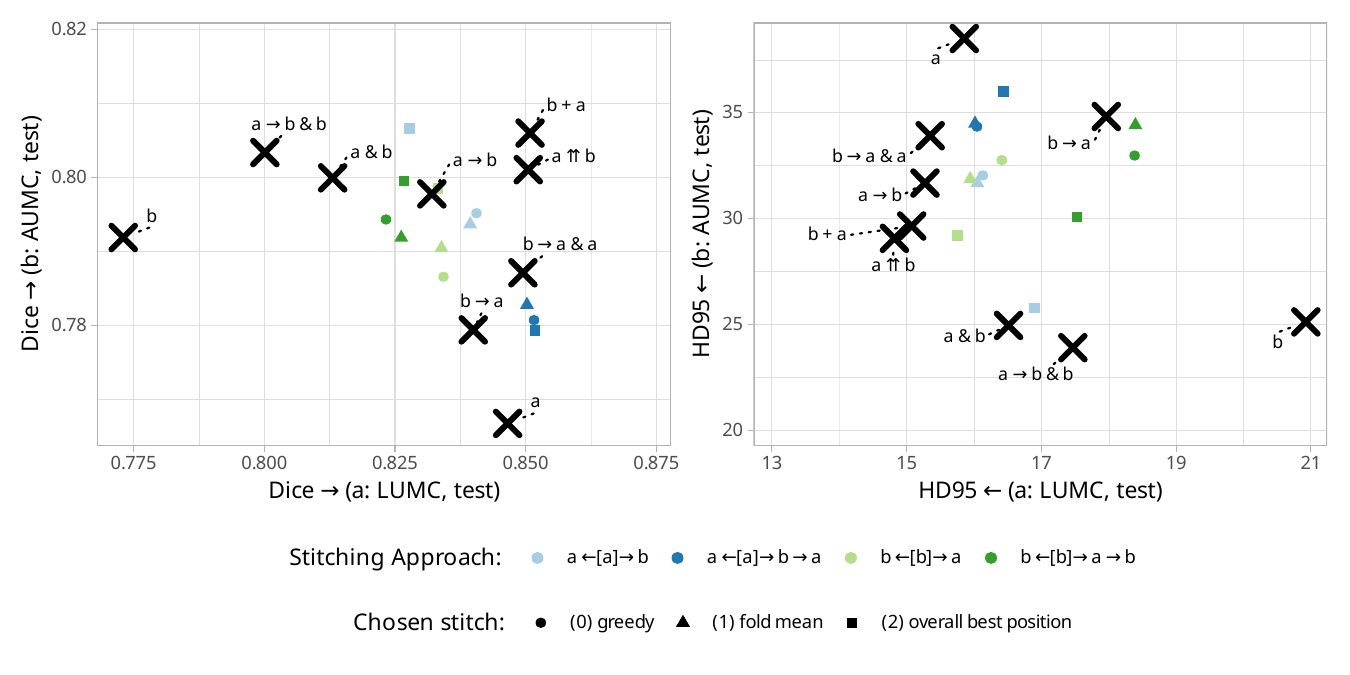}%
    \vspace{-0.7em}%
    }\\%
    \subcaptionbox{%
        Performance for dataset pair (a) HyperKvasir \& (b) CVC-ClinicDB
    }{%
    \includegraphics[scale=0.67]{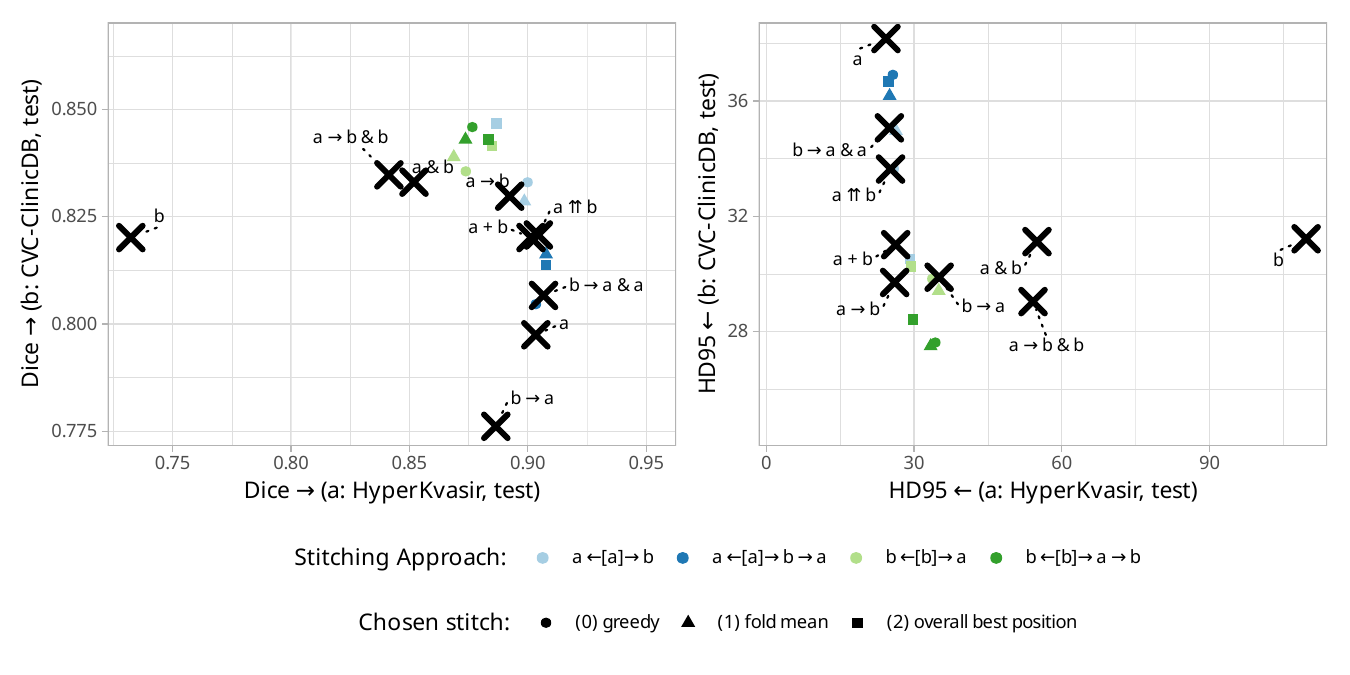}%
    \vspace{-0.7em}%
    }
    \caption{Mean performance of baselines and selected stitches on test set.}
    \label{fig:stitch-test-performance}
\end{figure}

\paragraph{Stitch Selection \& Stitch Performance}
In \cref{fig:stitch-test-performance} the performance of selected stitches on the test set is shown. The use of stitches can provide a favorable trade-off compared to the alternate approaches without having to share data, perform online training, and the requirement of fine-tuning with labelled data. The impact of choice of stitch varies. In particular, the rule of thumb is notably different in its tradeoff from greedy and fold-mean methods in a number of instances.

\begin{figure}
    \centering%
    \includegraphics[scale=0.5]{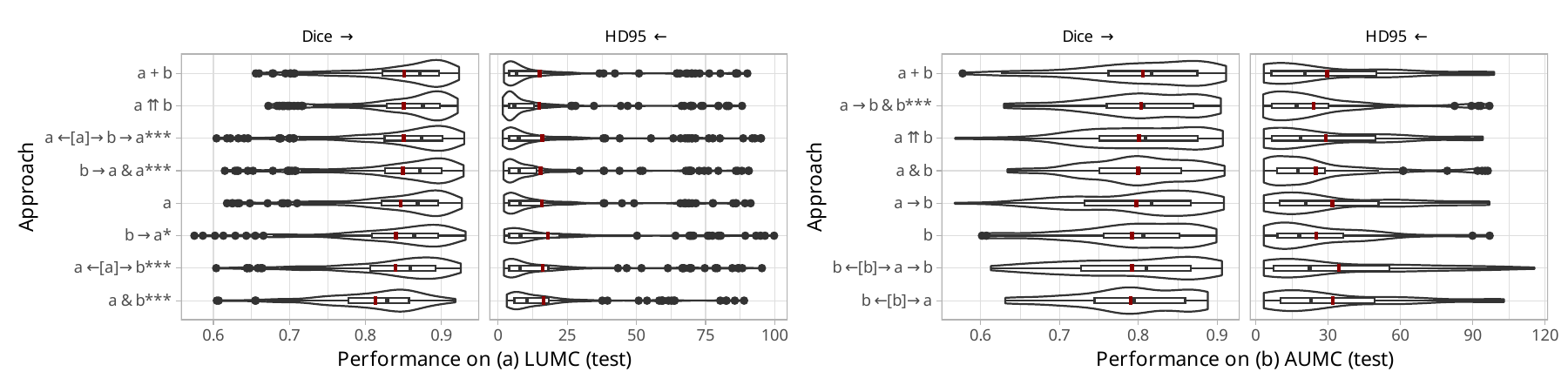}%
    \caption{Performance by approach, as combined box-plot violin-plot, for the viewpoint for LUMC (left) and AUMC (right). Red line represents the mean performance. Stitching position is chosen by \cref{stitch-sel:accumulate}. Significance in means for Dice with respect to training on own data alone (approach type `a' or `b') is indicated with \texttt{*}, for $P \leq 0.05$, \texttt{**}, for $P \leq 0.01$, \texttt{***}, for $P \leq 0.001$. Approaches are in order of mean performance.}%
    \label{fig:per-patient-own-data-performance-lumc-aumc}
\end{figure}

\begin{figure}
    \centering%
    \includegraphics[scale=0.5]{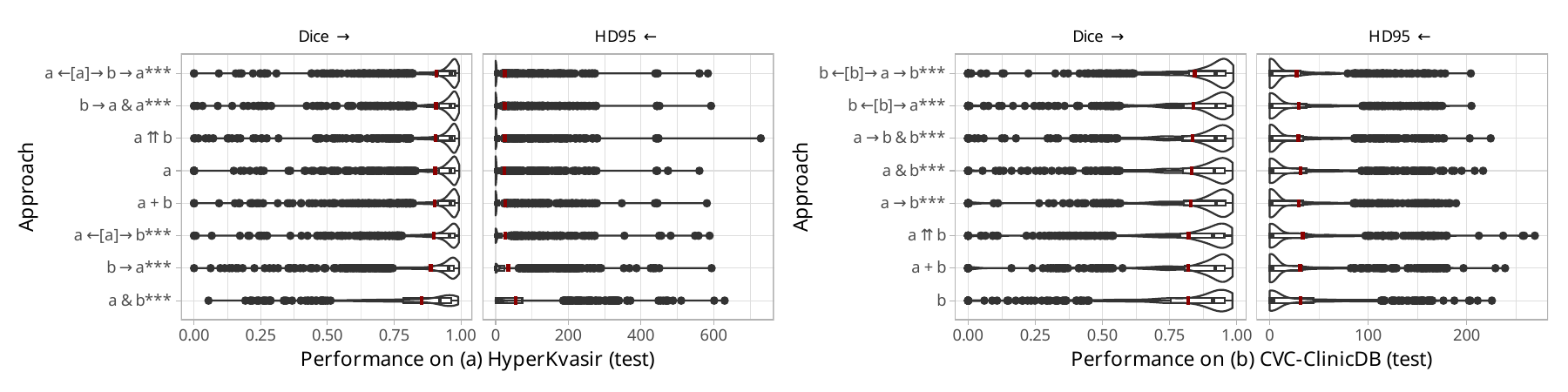}%
    \caption{Performance by approach, as combined box-plot violin-plot, for the viewpoint of HyperKvasir (left) and CVC-ClinicDB (right). Red line represents the mean performance. Stitching position is chosen by \cref{stitch-sel:accumulate}. Significance in means for Dice with respect to training on own data alone (approach type `a' or `b') is indicated with \texttt{*}, for $P \leq 0.05$, \texttt{**}, for $P \leq 0.01$, \texttt{***}, for $P \leq 0.001$. Approaches are in order of mean performance.}%
    \label{fig:per-patient-own-data-performance-hyperkvasir-cvcclinicdb}
\end{figure}

\paragraph{Own Data Performance Comparison - Do models trained with more varied data perform better on own data?}
Within this work we have investigated a variety of approaches that allow for collaboration between two parties with local datasets. In practice, performance on ones' own data is the most important. The results of this comparison can be found in \cref{fig:per-patient-own-data-performance-lumc-aumc,fig:per-patient-own-data-performance-hyperkvasir-cvcclinicdb}. For this comparison, we again only consider scenarios in which access to the dataset used is plausible, for example, we cannot fine-tune a network using someone else's local data ourselves. 

Overall, the distributions across various approaches tend to be highly similar, with some exceptions. We note that performance is strongly correlated with the specific sample used. However, in most instances differences are minimal, even if statistically significant.

On LUMC and AUMC data, the HD95 is low for most patients, indicating that most errors occur close to the ground-truth label in regions where inter-observer variation is known to be prevalent. Outliers do occur and often correspond to single-slice errors, which may be partly due to the use of 2D models rather than 3D models. For the HyperKvasir and CVC-ClinicDB datasets, HD95 correlates strongly with the Dice score. Both datasets include samples on which no model makes a nearly correct prediction, though this effect is more pronounced on the CVC-ClinicDB dataset, as performance on frames from the same video is heavily correlated.

As noted previously, the ensemble of the basis networks (basic ensemble) -- the networks without any fine-tuning -- performs badly on the test-set of the larger party of the two. From these figures it is apparent that this combination degrades performance generally, across all samples, from the perspective of a party looking at performance on their own data. Despite the favorable requirements (model shared once after training, no architecture requirements) as a model combination technique, and increased robustness to dataset shifts, these results make the basic ensemble less usable for the party with the larger dataset. Other techniques maintain performance much better compared to a model trained solely on local data. This includes stitching, which closes the gap with the addition of a stitch that only needs to be trained using (unlabeled) data.
In general, any major gains in performance from combining and sharing data are to be found in the robustness of the resulting models with regard to data shifts.
%

\section{Discussion}
In this work, we provided an improved procedure for training stitches, which improves the robustness of stitches when many stitches are used (answering \cref{RQ:construct-better-network}). In general, being able to combine datasets, or perform federated learning, provides robust networks, but any combination results in a more robust approach compared to training on a single dataset (answering \cref{RQ:comparison-combination}).

Further, we have shown that the addition of a single stitch to an ensemble can improve performance and can close the gap between the basic ensemble and the best performing methods (answering \cref{RQ:impact-stitching}). A neural network is more than just its output predictions, and intermediate layers can be highly informative. Stitching can be a highly useful technique that does not require data sharing or online collaboration. Furthermore, unlike fine-tuning, it does not necessarily require ground truth labels, and provides a simple tool to improve an ensemble.

We have limited ourselves to simple binary (foreground / background) segmentation tasks. This was done to avoid averaging metrics over multiple regions of interest. In practice, each region of interest is important, and performance degradation in each part is undesirable. At the additional cost of having a network for each region of interest, one can select a stitch independently for each region of interest, as was done within this work for additional flexibility.

However, stitching as explained here may lead to greater complexity than necessary. As a stitched network is effectively an enhanced ensemble, the computational cost is similar to that of an ensemble with the additional overhead of the stitches. It may be worthwhile to, for example, distill the knowledge~\autocite{hintonDistillingKnowledgeNeural2015} of the stitched network into a more compact format.

\paragraph{Multi-Task Learning} Finally, in this work we have combined networks trained on datasets for the same task, and assume the underlying definitions of each label are consistent. In practice, this may not always be the case, as different protocols may be used at different parties, causing discrepancies in ground truth. Unlike stitching with a task-specific loss~\cite{bansalRevisitingModelStitching2021,panStitchableNeuralNetworks2023} the method for training stitches presented here is not problem dependent. While this means we may no longer be able to create an ensemble as  done in this work, combining using stitching would still be possible.

%


\section{Conclusion}
We have investigated various approaches for creating a joint model when parties are unable to publish or share data. First, we demonstrated that creating a joint model offers added value compared to training on a single party's data. The diversity of data used to train a neural network directly influences how well the model generalizes, and combining models can enhance this generalization. For instance, a network trained on data that lacks diversity, may perform significantly worse when encountering data shifts. Any reasonable method for combining data or networks provides some robustness against such shifts. We have shown that stitching can improve an ensemble, significantly closing the gap between online collaboration and data sharing methods.


\section*{Acknowledgements}
This publication is part of the project "DAEDALUS - Distributed and Automated Evolutionary Deep Architecture Learning with Unprecedented Scalability" with project number 18373 of the research programme Open Technology Programme which is (partly) financed by the Dutch Research Council (NWO). Other financial contributions as part of this project have been provided by Elekta AB and Ortec Logiqcare B.V..

We thank Monika Grewal, Vangelis Kostoulas, and Alexander Chebykin for providing assistance with the data processing.

\printbibliography[]

\appendix

\section{Matching Layers in Neural Networks}\label{appendix:matching}
In the main body of the article we discuss the matching of layers in Neural Networks for stitching. This matching procedure is  a key component of the approach in \autocite{guijtExploringSearchSpace2024}, whose role it is to match layers between the two networks, such that no cycles are created. The work itself already notes that the matching is a weak-point of the work, with large gaps between some of the stitches. Therefore, we revisit the acyclic matching approach.

Unlike the architectures stitched in \autocite{guijtExploringSearchSpace2024}, the architectures within this work are both constructed by nnUNet, following a similar template. The resulting matchings from the approach, are hence trivial enough that they could be designed manually. Making the improvement less relevant to the article itself. However, we have found that this method is capable of finding matchings with more matches, faster.


\paragraph{Metric} A key issue with the original approach lies in the distribution of matches between the networks. With the approach from \autocite{guijtExploringSearchSpace2024}, a large portion of the layers has no stitches in between. As proposed in \autocite{panStitchableNeuralNetworks2023}, if a network twice as deep, for the deeper network a match every other layer would be preferred, as this reduces the impact of a single stitch.

To enable for an acyclic directed graph, we propose to score matches such that their distribution is improved, prior to running a matching algorithm.   

To quantify this, we define the progress along the network $\mathrm{p}$ for a layer $v$ as follows:
\begin{subequations}
    \begin{equation}
        \mathrm{p}(v) = \frac{\mathrm{d}_\mathrm{max}(v_\mathrm{in}, v)}{\mathrm{d}_\mathrm{max}(v_\mathrm{in}, v) + \mathrm{d}_\mathrm{max}(v, v_\mathrm{out})},
        \label{eqn:progress-relative-distance}
    \end{equation}
    
    The similarity between $v_a$, a layer of network $a$, and $v_b$, a layer of network $b$, is then taken as the squared difference between the progress of the two layers, if matching these two layers is possible:
    
    \begin{equation}
        \textrm{S}_\textrm{P}(v_a, v_b) = \begin{cases}
            1 - c\,(\mathrm{p}(v_a) - \mathrm{p}(v_b))^2 & \text{ if suitable stitching layer exists} \\
            -1 & \text{ otherwise}
        \end{cases}, \label{eqn:similarity-relative-distance}
    \end{equation}
\end{subequations}
where $c \in [0, 1]$ is a constant that can interpolate between the original count based matching function. This similarity can be used to calculate an initial similarity matrix $\textrm{S}_\textrm{P}$, between layers in network $a$ and layers in network $b$.

Beyond positional similarity, parallel branches may exist. For a computational cost reduction to be obtained, parallel branches need to be accounted for. As layers with high input or output cardinality are the starting and ending points of these parallel branches, ensuring these layers are matched is key to reducing the size of blocks. Furthermore, matching these points to each other is a reasonable heuristic, e.g., by matching the end of a block to the end of another block. To do so, we can adjust the similarity matrix. Given the input degree $d^\mathrm{in}_{v}$ and output degree $d^\mathrm{out}_{v}$ of a layer $v$:

\begin{equation}
    \textrm{S}_\textrm{PC}(v_a, v_b)  = \begin{cases}
        \textrm{S}_\textrm{P}(v_a, v_b) + d^\mathrm{in}_{v_a} d^\mathrm{in}_{v_b} + d^\mathrm{out}_{v_a} d^\mathrm{out}_{v_b} & \text{ if suitable stitching layer exists} \\
        -1 & \text{ otherwise}
    \end{cases}. \label{eqn:similarity-include-cardinality}
\end{equation}

\paragraph{Acyclic Matching} For the resulting similarity matrix $\textrm{S}$, pairs of layers between the two networks need to be selected, such that no cycle is constructed. If both networks were sequential, this problem could be solved using dynamic programming. Here, this is performed using a modified version of Hirschberg's Algorithm~\autocite{hirschbergLinearSpaceAlgorithm1975}, an algorithm for computing the edit distance between two strings. Instead of adding $1$ if the 'character' is the same, the similarity matrix is used instead to score matching a pair of layers.

This approach can also be used for acyclic graphs. The network represented by an acyclic graph needs to execute in some ordering. If one fixes these orderings in place, the aforementioned approach can again be applied. In exchange for losing the guarantee of finding the optimum over all possible topological orderings, the resulting approach is much faster. Within this work we use only the default topological ordering. This ordering corresponds to which vertices were introduced first. We note that this ordering is equal to the ordering in which the original Python code used these layers, the calling order.

As networks used within originate from the same parameterized architecture provided by nnUNet, the calling order is consistent between networks. Hence, using this ordering, excluding the architectural differences between the networks, the current procedure already results in a one-to-one match.



\section{Improving Training}\label{appendix:double-batched-training}
Within this work we propose another method for training stitches that lies in-between direct matching, and task-loss matching. Direct matching provides a simple method that only accounts for the feature map at a given point, allowing all stitches to be trained independently in parallel using the same forward and backward pass. Task-loss matching allows for improvements to the feature representation by the impact of later layers, but does not allow for this parallelization, and, on its own, tends to create out-of-distribution activations.

To improve upon, and combine the strengths of these two prior approaches to training stitches, we propose a new approach: \emph{double-batched training}. Unlike direct matching, we can account for the most impactful errors in the stitch on later feature maps, and unlike task-loss we remain able to train multiple stitches in parallel.
When training stitches using direct matching, at each switch after the stitch we have access to two feature maps, is the stitched feature map prediction, and the original (reference) input. During normal training, the switch will return the original input, as to maintain this reference feature map for training. In turn, the stitched feature map is dropped.

Double-batched training preserves this feature map by using the batch-dimension not just for independent samples, but also for preserving the stitched feature map. It accomplishes this by doubling the batch size of any input that goes into the neural network, initially by providing each sample twice. 

\begin{figure}[tbhp]
    \centering
    \includegraphics[width=0.7\columnwidth]{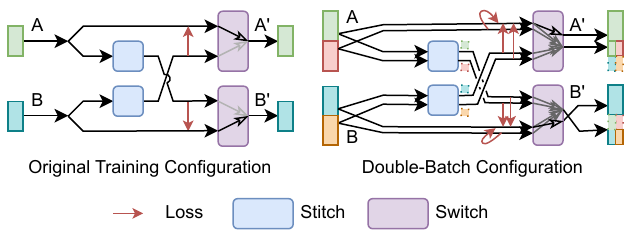}%
    \caption{%
        The layers that have been matched have corresponding feature maps $A$ and $B$ as output. Within the original approach (left) the original values (lime, teal) have been provided and need to be forwarded such that layer stitches may be trained using the original reference value. By using the batch dimension to potentially store another variant of the feature map, for example one resulting from a current or previous stitch, the influence of layers can be assessed by computing a loss at the next stitch or intermediate layers.%
        \label{fig:old-and-new-training-configuration}%
    }%
\end{figure}

This is schematically visualized in \cref{fig:old-and-new-training-configuration}, where box represents a possibly stitched variant of feature map of an individual sample. The first (1, green \& teal) position in each batch always corresponds to the feature map without any stitches, whereas the second (2, red and orange) of the batch can be any feature map, including those involving a previous stitch.

In turn, at the switch (purple) we have access to four variants of what should correspond to the same feature map. (1) original, (2) forwarded, (3) stitched original (from other network), (4) stitched forwarded (from other network). Losses are computed between (1) and (2), (3) and (4). The output of the switch maintains the invariant that the first value maintains the original as a reference value, and randomly selects any of the feature maps at random.

This new approach allows for the impact of a stitch to be calculated at later points in the network, allowing stitches to adapt to the impact of later layers, including other stitches. Furthermore, all stitches are still involved with at least one loss providing a gradient during each forward and backward pass. 

However, these benefits come at a computational cost. As the same sample is provided twice, to maintain the same batch size, only half the unique samples are present per batch. Much like other approaches that propagate stitched feature maps through the network, we need to ensure that any layers that compute batch statistics do not update their statistics during training with this method. Furthermore, gradients will need to be back-propagated through the network, where this previously was not necessary.

\end{document}